\documentclass[journal]{IEEEtran}
\IEEEoverridecommandlockouts
\usepackage[utf8]{inputenc}
\usepackage{newunicodechar}
\newunicodechar{−}{-}
\usepackage{hyperref}
\usepackage{amsmath,amssymb,amsfonts}
\usepackage{graphicx}
\usepackage{textcomp}
\usepackage[table]{xcolor}
\usepackage{subcaption}
\usepackage{gensymb}
\usepackage{float}
\usepackage{algorithm,algcompatible,algpseudocode}
\usepackage{array}
\usepackage{eso-pic, rotating, graphicx}
\usepackage{multirow}
\usepackage{multicol}
\usepackage{tabularx, makecell}
\usepackage{hyperref}
\usepackage[utf8]{inputenc}
\usepackage{svg}
\usepackage{ctable}
\usepackage{booktabs}
\usepackage{lineno}
\usepackage{threeparttable}
\usepackage[justification=centering]{caption}
\usepackage{placeins}  
\usepackage{comment}
\usepackage[acronym]{glossaries}
\makeglossaries
\newif\ifclean
\cleantrue
\ifclean\colorlet{blue}{black}\fi
\newcommand{\rev}[1]{\ifclean#1\else{\color{blue}#1}\fi}
\newcommand{\revsec}[1]{\texorpdfstring{\rev{#1}}{#1}}
\newcommand{\revg}[1]{\ifclean#1\else{\color{blue}#1}\fi}
\newcommand{\revgsec}[1]{\texorpdfstring{\revg{#1}}{#1}}
\newacronym{ai}{AI}{Artificial Intelligence}
\newacronym{gnss}{GNSS}{Global Navigation Satellite Systems}
\newacronym{dos}{DoS}{Denial-of-Service}
\newacronym{heterogclstm}{HeteroGCLSTM}{Heterogeneous Graph Convolutional Long Short-Term Memory}
\newacronym{gcnn}{GCNN}{Graph Convolution Neural Network}
\newacronym{cnn}{CNN}{Convolutional Neural Network}
\newacronym{dtdg}{DTDG}{ Discrete-Time Dynamic Graphs}
\newacronym{ctdg}{CTDG}{ Continuous-Time Dynamic Graphs}
\newacronym{snr}{SNR}{Signal-to-Noise ratio}
\newacronym{vmd}{VMD}{Variational Mode Decomposition}
\newacronym{nlos}{NLOS}{Non-Line-of-Sight}
\newacronym{gnn}{GNN}{Graph Neural Networks}
\newacronym{gin}{GIN}{Graph Isomorphism Network}
\newacronym{jade}{JADE-TG}{Jamming-Aware Deviation Estimation via Temporal Graphs}
\newacronym{pec}{PEC}{Position error correction}
\newacronym{sdr}{SDR}{Software-Defined Radios}
\newacronym{std}{STD}{standard deviation}
\newacronym{mae}{MAE}{Mean Absolute Error}
\newacronym{jaguard}{JaGNet}{Jamming Graph Network}
\begin{document}

%\title{Deep Temporal Graph Networks for Correction of GNSS Jamming-Induced Deviations}
%\title{JaGNet: Jamming Correction of GNSS Deviation with Deep Temporal Graphs}

\title{JaGNet: Detection, Identification and Position Error Correction of GNSS Jamming with Deep Temporal Graphs}

\author{\IEEEauthorblockN{%
    Ivana Kesi\'c\IEEEauthorrefmark{1}\revg{\IEEEauthorrefmark{3}}, Alja\v{z} Blatnik\IEEEauthorrefmark{2}\revg{\IEEEauthorrefmark{3}}, Carolina Fortuna\IEEEauthorrefmark{1} and Bla\v{z}~Bertalani\v{c}\IEEEauthorrefmark{1} \\
}\IEEEauthorblockA{%
    \IEEEauthorrefmark{1}Department of Communication Systems, Jo\v{z}ef Stefan Institute, Slovenia\\
    \IEEEauthorrefmark{2}Faculty of Electrical Engineering, University of Ljubljana, Slovenia\\
    \revg{\IEEEauthorrefmark{3}These authors contributed equally.}\\
}

\{carolina.fortuna, blaz.bertalanic\}[at]ijs.si%
\\
}

\maketitle

\begin{abstract}

\gls{gnss} underpin location-dependent mobile computing, such as \rev{railway signaling, public transport and} fleet tracking, yet face growing disruption from low-cost jammers. \rev{ Under moderate jamming, low-cost receivers keep reporting a valid fix, shifted by up to several meters and often without an interference flag. Their built-in defenses run in closed firmware, and existing data-driven position error correction methods are designed for multipath, require raw pseudoranges or operate on single epochs.
We propose JaGNet, a single-layer heterogeneous graph convolutional LSTM over star graphs of the receiver and its satellites, which, from the standard receiver output alone, detects the jamming, identifies its type and power, and corrects the reported position only when jamming is detected. We evaluate it on new hardware-in-the-loop recordings of a commercial multi-constellation receiver under three jammer types at six powers, at six static locations and on eleven railway and bus routes. JaGNet detects 99.8 to 99.9\% of the jammed 10-s windows with at most 1.0\% false positives, against 64.7 to 66.2\% for the receiver's own monitor, and identifies the jammer type in 95.2 to 96.6\% of them. Trained in the laboratory only, it detects 97.7\% of the Jammertest 2024 field windows with a $C/N_0$ drop of 10 dB or more. It reduces the mean absolute position error by 67.5\% at held-out locations and by 64.0\% on held-out route repetitions, where its error is 20.2 cm below that of the strongest baseline, whereas at the static locations the attention-based baselines are within 0.4 cm of it.}
\end{abstract}

\glsresetall % resetting after \chapter works
% \begin{IEEEkeywords}
% GNSS jamming, position error correction, dynamic graph neural networks, network resilience, satellite networks, adversarial network degradation
% \end{IEEEkeywords}
\section{Introduction}
%What is important? (1-2 paragraphs)
%What is missing? (1-2 paragraphs)
%How do we solve it? 

\textit{Motivation:} \gls{gnss} satellite constellations constitute one of the most pervasive engineered wireless networks in operation, providing positioning and timing services to billions of devices across autonomous driving~\cite{xiong2021g}, connected vehicles~\cite{mahmoud2020integrated}, \rev{railway signaling~\cite{marais2017survey},} aviation~\cite{workgroup2024gps}, power-grid synchronization~\cite{biswas2024cyber} and  telecommunications~\cite{li20235g}. Like any wireless network, this infrastructure is vulnerable to adversarial degradation. Civil aviation alone recorded a five-fold increase in large-scale disruptions in 2024~\cite{workgroup2024gps}. Intentional jamming degrades the satellite--receiver network by suppressing link quality and forcing topology changes: satellites drop from the receiver's view as tracking is lost, severing edges in the network graph. The resulting positional drift of up to several meters directly threatens the integrity thresholds mandated by standards bodies~\cite{5GAA2021V2XHAP}. \revg{In railway signaling, for example, the position decides when a train passes a virtual balise~\cite{azaola2018virtual}, and the European Union Agency for the Space Programme (EUSPA) requires at most 2.5 m on each axis at 95\% of the time for the absolute train position~\cite{euspa2025rail}.} This has motivated the development of resilient frameworks that address adversarial interference through dynamic resource allocation~\cite{shao2024deep} or spatially encoded signaling~\cite{reusmuns2023flying}, highlighting a broader shift towards integrating software-defined robustness into mobile platforms that lack specialized hardware countermeasures.

\begin{figure*}[!t]
\centering
\includegraphics[width=\textwidth]{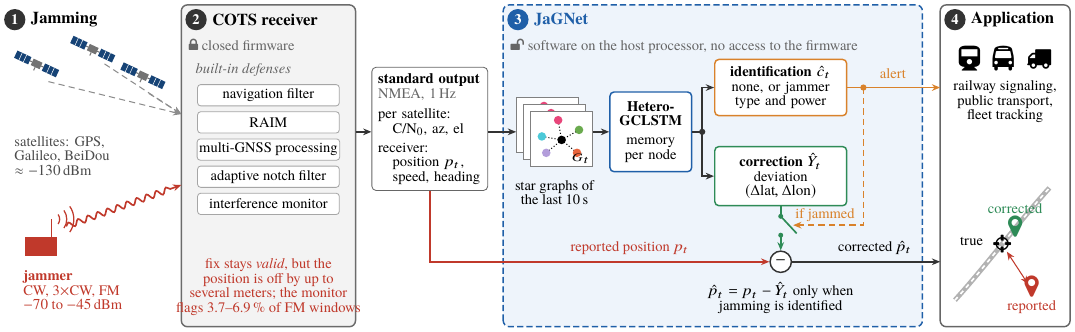}
\caption{\revg{\acrshort{jaguard} as a software layer between a low-cost \gls{gnss} receiver and the application. Under jamming (1), the receiver keeps reporting a valid but shifted fix (2). From the standard output alone, \acrshort{jaguard} identifies the jamming and corrects the position (3) for the application (4).}}
\label{fig:cots}
\end{figure*}

The fundamental vulnerability stems from the extremely weak received \gls{gnss} signal power, further degraded by satellite aging~\cite{bermudez_10947330}. Even modest interference can suppress the effective \gls{snr} and destabilize tracking loops. Jamming remains the most operationally disruptive mechanism~\cite{morong2019study}, as low-cost \gls{sdr}s now enable agile, programmable jammers that are easy to deploy~\cite{garcia2025characterization}. This threat is acute for the popular receivers, such as U-blox MAX-M10, as these low-cost units lack the hardware countermeasures of survey-grade equipment~\cite{hamza2025recent}. Although receivers incorporate spread-spectrum anti-jamming measures~\cite{gerrard2022exploration}, contemporary threats increasingly exceed the capabilities of traditional filtering and blanking~\cite{borio2021interference}. \rev{\label{rev:R2.3-intro}These built-in defenses, such as navigation filtering, integrity monitoring, multi-constellation processing and adaptive notch filtering~\cite{kaplan2017gnss, moralesferre2020survey}, run inside closed firmware (Fig.~\ref{fig:cots}, stage~2). Notch filters suppress narrow tones, and integrity monitoring assumes that only a minority of the measurements is faulty, whereas jamming degrades all signals in its band at once~\cite{borio2016jammers}. \label{rev:R2.5-intro}\revg{Strong jamming causes a loss of fix, which the dependent application detects~\cite{kaplan2017gnss}, whereas moderate jamming shifts the reported position by one to several meters while the receiver keeps reporting a valid fix~\cite{blatnik2025evaluating}. An application that sees only the receiver output can correct this residual error only in the position domain}~\cite{park2013dgpscp}.}

\revg{\textit{Prior art:} Data-driven interference mitigation addresses anomaly detection~\cite{li2024novel, liu2023nlos}, position forecasting~\cite{zhou2025hybrid} and \gls{pec}~\cite{kanhere2022improving}. Anomaly detection identifies or locates the interference without correcting the position, and forecasting becomes unstable under non-stationary interference~\cite{lv2025review}. Existing \gls{pec} methods are designed for multipath rather than jamming~\cite{kanhere2022improving, zhu2024enhancing, liu2024deepgps}, operate on single epochs~\cite{mohanty2023learning}, or require raw pseudoranges, which the NMEA output of commercial receivers lacks~\cite{kanhere2022improving, weng2024prnet}. Temporal models over the constellation graph address only multipath~\cite{mohanty2024tightly, zhao2023gnnrl}, and, to our knowledge, none represents the receiver as a node. No open jamming dataset has been used for \gls{pec}, and none exposes one receiver to the same controlled interference at rest and in motion.}

\revg{\textit{Proposed approach:} In the absence of interference, the topology and link quality of the satellite constellation evolve continuously and predictably~\cite{shen2024factor}. Jamming disrupts this coherence: satellites are intermittently lost and reacquired, their \gls{snr} decreases asynchronously, and the receiver-satellite topology may change  irregularly, which a dynamic graph can model. We therefore propose \gls{jaguard}, a receiver-centric temporal graph network that detects the jamming, identifies its type and power, and estimates the horizontal position deviation (Fig.~\ref{fig:cots}, stage~3). \label{rev:R3.2-intro}Because learning the correction requires recordings with a known true position, it applies to receivers at fixed locations or on known routes, such as railway lines, whose track is surveyed~\cite{neri2018track, filip2017certification}, and bus lines~\cite{weng2024prnet}. The contributions of the paper are:}

\begin{table*}[!htbp]
    \centering
    \scalebox{1.0}{
    \begin{threeparttable}[b]
    \caption{Summary of related work in data-driven GNSS mitigation and graph learning methods in network science.}
        \begin{tabular}{l l l c c c}
            \toprule
            \textbf{Work} & \textbf{Goal} & \textbf{Method Category} & \textbf{Model} & \textbf{Input} & \textbf{Data} \\
            \midrule
        \multicolumn{6}{c}{\textit{Data-driven mitigation of adversarial GNSS network degradation}} \\
        \midrule

        % Anomaly Detection / Classification
        Liu~\textit{et al.}~\cite{liu2023nlos} & AD & Signal Processing + DL & VMD + CNN & MTS & Prop. \\
        Zhong~\textit{et al.}~\cite{zhong2024tsfanet} & AD & Deep Learning & CNN + RNN + Attention & Pre-correlation Spectrogram & Sim. \\
        Mehr~\textit{et al.}~\cite{mehr2024deep} & AD & Deep Learning & LSTM-Autoencoder + CNN & MTS & Sim. \\
        Spanghero~\textit{et al.}~\cite{spanghero2025gnss} & AD & Signal Processing & Spectral Correlation & Raw IQ & Prop. \\
        Wu~\cite{wu2024gnss} & AD & Signal Processing & RO Radiometry & RO SNR & Open \\
        \rev{Liu~\textit{et al.}~\cite{liu2022balise}} & \rev{AD} & \rev{Signal Processing} & \rev{PR Residual Test} & \rev{PR + Odometry + Track Map} & \rev{Prop.} \\
        \rev{Cao~\textit{et al.}~\cite{cao2024railway}} & \rev{AD} & \rev{Sensor Fusion} & \rev{Robust GNSS/INS Filter} & \rev{PR + INS + Odometer + Track Map} & \rev{Prop.} \\
        \midrule

        % Position Forecasting (Absolute)
        Zhou~\textit{et al.}~\cite{zhou2025hybrid} & PF & Signal Processing + DL & EEMD + LSTM & UTS & Open \\
        Wang~\textit{et al.}~\cite{liu2023vehicle} & PF & Deep Learning & 3D CNN-LSTM & MTS + Spatial Grids & Open \\
        Gao~\textit{et al.}~\cite{gao2022modelling} & PF & Supervised ML & GBDT, LSTM, SVM & MTS + External Features & Open \\

        \midrule
        % Position Correction (Error Regression)
        Kanhere~\textit{et al.}~\cite{kanhere2022improving} & PEC-MP & Deep Learning & Set-based DNN & Pseudorange Residuals, LOS & Open \\
        \rev{Zhao~\textit{et al.}~\cite{zhao2023gnnrl}} & \rev{PEC-MP} & \rev{Deep RL} & \rev{GNN + Recurrent RL} & \rev{Graph (Satellite Features)} & \rev{Open, Prop.} \\
        \rev{Weng~\textit{et al.}~\cite{weng2024prnet}} & \rev{PEC-MP} & \rev{Deep Learning} & \rev{PrNet (MLP per satellite)} & \rev{C/N$_0$, El., WLS Position, Geometry} & \rev{Open} \\
        Zhu~\textit{et al.}~\cite{zhu2024enhancing} & PEC-MP & Deep Learning & Transformer & MTS & Open \\
        Mohanty~\textit{et al.}~\cite{mohanty2023learning} & PEC-MP & Deep Learning & GCNN & Graph (Satellite Features) & Open \\
        Liu~\textit{et al.}~\cite{liu2024deepgps} & PEC-MP & Deep Learning & 2D CNN Encoder-Decoder & MTS + Map & Prop. \\

        \midrule
        \multicolumn{6}{c}{\textit{Graph learning methods for network-level tasks}} \\
        \midrule
        Skarding~\textit{et al.}~\cite{skarding2023effectiveness} & Link Pred. & DL + Heuristics & GNN Ensemble & Dyn. Graph & Open \\
        Wang~\textit{et al.}~\cite{wang2023timesensitive} & Net. Sched. & Graph + Optim. & CQF Sched. & Sat. Topology & Sim. \\
        Huang~\textit{et al.}~\cite{huang2024routing} & Net. Routing & Graph + Optim. & ILA + QoSR & Sat. Topology & Sim. \\
        Li~\textit{et al.}~\cite{li2024trajectory} & Traj. Pred. & Deep Learning & Graph ST & Dyn. Graph & Open \\
        Zong~\textit{et al.}~\cite{zong2025resilient} & Net. Resil. & Deep Learning & Dist. GNN & Dyn. Graph & Sim. \\

        \midrule
        \textbf{This work} & \textbf{PEC-J} \rev{\textbf{+ AD}} & \textbf{Deep Learning} & \textbf{Temporal Graph NN} & \textbf{DHG} & \textbf{Open}\rev{$^\dagger$} \\
        \bottomrule
    \end{tabular}
    \label{tab:related}
    \begin{tablenotes}
        \item[*] AD (Anomaly Detection), PF (Position Forecasting), PEC-MP (Position Error Correction - MultiPath), PEC-J (Position Error Correction - Jamming), Link Pred. (Link Prediction), Net. Sched. (Network Scheduling), Net. Routing (Network Routing), Traj. Pred. (Trajectory Prediction), Net. Resil. (Network Resilience), UTS (Univariate Time Series), MTS (Multivariate Time Series), Map (City Environment Map), DHG (Dynamic Heterogeneous Graph), Dyn. Graph (Dynamic Graph), Sat. Topology (Satellite Network Topology), Graph ST (Graph Spatio-Temporal), CQF Sched. (Cyclic Queuing and Forwarding Scheduling), ILA + QoSR (Inter-Layer Link Allocation + QoS-based Routing), Dist. GNN (Distributed GNN), DL (Deep Learning), LOS (Line-of-Sight), RO (Radio Occultation)\rev{, PR (Pseudorange), INS (Inertial Navigation System), WLS (Weighted Least Squares)}. Data: Sim. (Simulated), Prop. (Proprietary), Open (Publicly Available)\rev{, $^\dagger$released upon acceptance}.
    \end{tablenotes}
    \end{threeparttable}
    }
\end{table*}

\begin{itemize}

    \item \revg{\textit{Formulation and model.} We formulate the detection, identification and correction of \gls{gnss} jamming as one learning problem on heterogeneous star graphs with the receiver as a node, and propose \gls{jaguard}, a single-layer HeteroGCLSTM with 69,205 parameters whose identification head gates its correction head. It requires only the standard output of low-cost receivers used in railway signaling, public transport and fleet tracking, and processes a 10-s window in 37.3 to 37.7\,ms on a Raspberry Pi~5.}

    \item \revg{\label{rev:R2.1-contrib}\textit{Dataset and evaluation protocol.} A \textit{new} hardware-in-the-loop dataset of a commercial multi-constellation receiver under three jammer types at six powers, at six static locations and on eleven railway and bus routes (1,615 recordings), to be released with the code upon acceptance\footnote{Code:\url{https://github.com/sensorlab/JaGuard/}}, and an evaluation on held-out locations and route repetitions with paired statistics and over-the-air detection tests.}

    \item \revg{\textit{Detection and identification.} \gls{jaguard} detects 99.8 to 99.9\% of the jammed windows with at most 1.0\% false positives, including 100.0\% of the FM-jammed windows, of which the receiver internal monitor flags only 3.7 to 6.9\%, and identifies the jammer type in 95.2 to 96.6\% of them. Trained in the laboratory only, it detects 97.7\% of the Jammertest 2024~\cite{sayyaf2025jammertest} windows with a C/N$_0$ drop of 10\,dB or more and all 28 events with a drop above 6\,dB on an Antarctic cruise, with at most 1.5\% false positives.}

    \item \revg{\textit{Correction.} The gated correction reduces the MAE of jammed fixes by 67.5\% at held-out static locations and by 64.0\% on held-out route repetitions, where its MAE is 20.2\,cm (95\% CI [16.4, 24.3]) below that of the strongest baseline, significantly on nine of the eleven routes. In the static scenario, MAE of the attention-based baselines is within 0.4\,cm of that of \gls{jaguard}. On a new route, ten calibration recordings raise the error reduction to 47.3 to 56.2\%, against 3.3 to 33.5\% without them.}
\end{itemize}

The remainder is organized as follows: Section~\ref{sec:related} reviews related work, Section~\ref{sec:prob:formulation} formulates the problem, Section~\ref{sec:transf} presents the proposed method, Section~\ref{sec:expsetup} describes the experimental setup, Section~\ref{sec:results} analyzes results, Section~\ref{sec:limitations} discusses limitations, and Section~\ref{sec:conclusions} concludes.

\section{Related Work}
\label{sec:related}

Our work connects two distinct research domains: data-driven \gls{gnss} interference mitigation and graph learning for network-level tasks. Data-driven \gls{gnss} mitigation methods focus on three objectives: (1) anomaly detection, (2) absolute position forecasting, and (3) direct position error correction. Meanwhile, graph neural networks have established themselves as a powerful tool for network-level tasks in engineered systems. Table~\ref{tab:related} is structured around these two strands, grouping related work into \textit{data-driven \gls{gnss} mitigation} and \textit{graph learning for network-level tasks}, and comparing each by goal, method category, model architecture, input type, and dataset availability.

\subsection{Data-driven \gls{gnss} interference mitigation}

A major portion of the \gls{gnss} mitigation literature\rev{~\cite{marata2026trends}} addresses \textit{anomaly detection}: identifying irregularities for retrospective correction~\cite{liu2023nlos} or event \revg{identification} and situational awareness~\cite{mehr2024deep, zhong2024tsfanet, spanghero2025gnss, wu2024gnss}. For example, Liu et al.~\cite{liu2023nlos} use a \gls{cnn} and \gls{vmd} to detect and correct \gls{nlos}-corrupted pseudoranges in smartphone \gls{gnss} data. For intentional interference, Mehr et al.~\cite{mehr2024deep} \revg{identify the} interference type from raw signals, Zhong et al.~\cite{zhong2024tsfanet} achieve high-accuracy jamming recognition from pre-correlation spectrograms, \rev{van der Merwe et al.~\cite{vandermerwe2023lowcost} detect and \revg{identify} interference with machine learning on the raw signal samples and measurements of a low-cost monitoring receiver built from commercial off-the-shelf components,} Spanghero et al.~\cite{spanghero2025gnss}  employ drones with horizon scanning and triangulation to geolocate jammers, and spectral correlation to identify jammer type, while Wu~\cite{wu2024gnss} monitors regional jamming from LEO satellites using Radio Occultation \gls{snr}. \rev{In railway signaling, Liu et al.~\cite{liu2022balise} predict the pseudoranges from odometry and the track map, detect jamming in the residuals and isolate the degraded measurements, while Cao et al.~\cite{cao2024railway} discriminate jamming within a tightly coupled GNSS/INS filter and compensate for the degraded measurements with the inertial sensor, the odometer and the track map. Commercial receivers also report an interference state of their own, such as the jamming indicator of u-blox receivers~\cite{ublox_m10_ifd}, which serves as a reference in Section~\ref{sec:res:ident}.} However, all these approaches are purely reactive: they identify or locate attacks\rev{, or compensate for them with other sensors,} but do not correct the corrupted position.

\textit{Position forecasting} methods predict absolute coordinates from historical trends. Wang et al.~\cite{liu2023vehicle} use a 3D \gls{cnn}-LSTM within particle filtering to predict vehicle position by jointly processing \gls{gnss} kinematic features and geographic layer information. Zhou et al.~\cite{zhou2025hybrid} apply EEMD signal decomposition followed by LSTM forecasting, while Gao et al.~\cite{gao2022modelling} augment time series with external physical variables, such as atmospheric pressure. However, forecasting absolute position is inherently unstable~\cite{lv2025review}: the non-stationary signal causes uncorrected historical errors to compound, leading to predictive drift.

\textit{\gls{pec}} instead predicts the deviation in longitude and latitude, reframing the task as a more stable regression. Kanhere et al.~\cite{kanhere2022improving} estimate 3D corrections from pseudorange residuals using a set-based DNN, \rev{Weng et al.~\cite{weng2024prnet} correct each pseudorange with a satellite-wise MLP before the position is solved, on routes represented in training,} Zhu et al.~\cite{zhu2024enhancing} employ a Transformer reformulating regression as classification, Mohanty et al.~\cite{mohanty2023learning} use a \gls{gnn} to learn corrections from satellite spatial relationships, and Liu et al.~\cite{liu2024deepgps} fuse building-height context with satellite statuses in an encoder-decoder network to correct GPS positions on mobile devices in urban canyons. \rev{Temporal extensions of graph-based correction filter per-epoch graph outputs with a Kalman filter~\cite{mohanty2024tightly} or pass per-epoch graph embeddings to a recurrent reinforcement-learning policy~\cite{zhao2023gnnrl}. Both construct the graph over the satellites only and target multipath.}

However, these \gls{pec} architectures are designed for multipath, not jamming. Kanhere et al.~\cite{kanhere2022improving}\rev{, Weng et al.~\cite{weng2024prnet}} and Mohanty et al.~\cite{mohanty2023learning} require raw pseudorange residuals, Doppler, and carrier-phase data unavailable in standard receiver settings, and assume a reliable initial position, which strong interference such as jamming biases or removes. Moreover, the satellite-mesh topology of Mohanty et al.~\cite{mohanty2023learning} captures inter-satellite geometric correlations but \rev{has no node for the receiver state, which carries the correction in motion (Section~\ref{sec:res:ablation})}.
Liu et al.~\cite{liu2024deepgps} require building-height and road-distribution maps as explicit input features to identify which satellites are geometrically obstructed. Without such city-map data, the model cannot operate, restricting deployment to urban zones with available map coverage. Furthermore, the model processes each epoch independently, without modeling the temporal evolution of the satellite network, limiting applicability to static geometric degradation rather than the time-correlated \gls{snr} loss and satellite outages induced by broadband jamming.

\subsection{Graph learning for wireless and mobile systems}
Graph neural networks have emerged as a powerful tool for wireless and mobile systems in engineered systems.  Mohanty et al.~\cite{mohanty2023learning} leverage them for \gls{pec} on multipath \gls{pec} dataset while Zong~\textit{et al.}~\cite{zong2025resilient} leverage distributed \gls{gnn}s for resilient clock synchronization in aerial swarms. These works represent the closest methodological parallel to our work, leveraging graph-based deep learning to achieve resilient estimation in a wireless network under adverse conditions. Dynamic \gls{gnn} architectures have been evaluated for temporal link prediction~\cite{skarding2023effectiveness}, and graph-based spatial-temporal models have been considered for trajectory prediction from evolving topologies~\cite{li2024trajectory}. In satellite networks specifically, dynamic topology modeling has been addressed for routing in multi-layer LEO/MEO constellations~\cite{huang2024routing} and time-sensitive scheduling~\cite{wang2023timesensitive}, establishing satellite constellations as an accepted network science domain.

\subsection{Gap Identification}
Existing works either target different problems or largely ignore the spatiotemporal structure of \gls{gnss} interference. Anomaly detection is purely reactive, forecasting amplifies errors once interference distorts the history, and \rev{the \gls{gnn}-based \gls{pec} methods process each epoch independently~\cite{mohanty2023learning} or filter per-epoch graph outputs over time~\cite{mohanty2024tightly, zhao2023gnnrl}, all for multipath rather than} the rapid disruption of spatiotemporal coherence caused by interference such as jamming. To the best of our knowledge, no prior work on network resilience has formulated \gls{gnss} jamming mitigation as a dynamic temporal graph regression problem. \gls{jaguard} addresses this gap by leveraging the receiver-centric star topology to jointly model evolving geometry and short-term history, capturing time-correlated distortions that static and set-based approaches overlook.

\section{Problem Formulation}
\label{sec:prob:formulation}

\gls{gnss} signals arrive below the receiver’s noise floor due to orbital free-space loss and spread-spectrum transmission. Receivers recover them by correlating the incoming signal with each satellite’s spreading code and integrating over short intervals, allowing the correlation peak to rise above the noise. Any additional radio energy in the L1-band ($\sim\!1575.42$\,MHz) acts as jamming noise, lowering the effective \gls{snr} and destabilizing satellite tracking. 

\begin{figure}[htbp]
  \centering
  \includegraphics[width=0.85\linewidth]{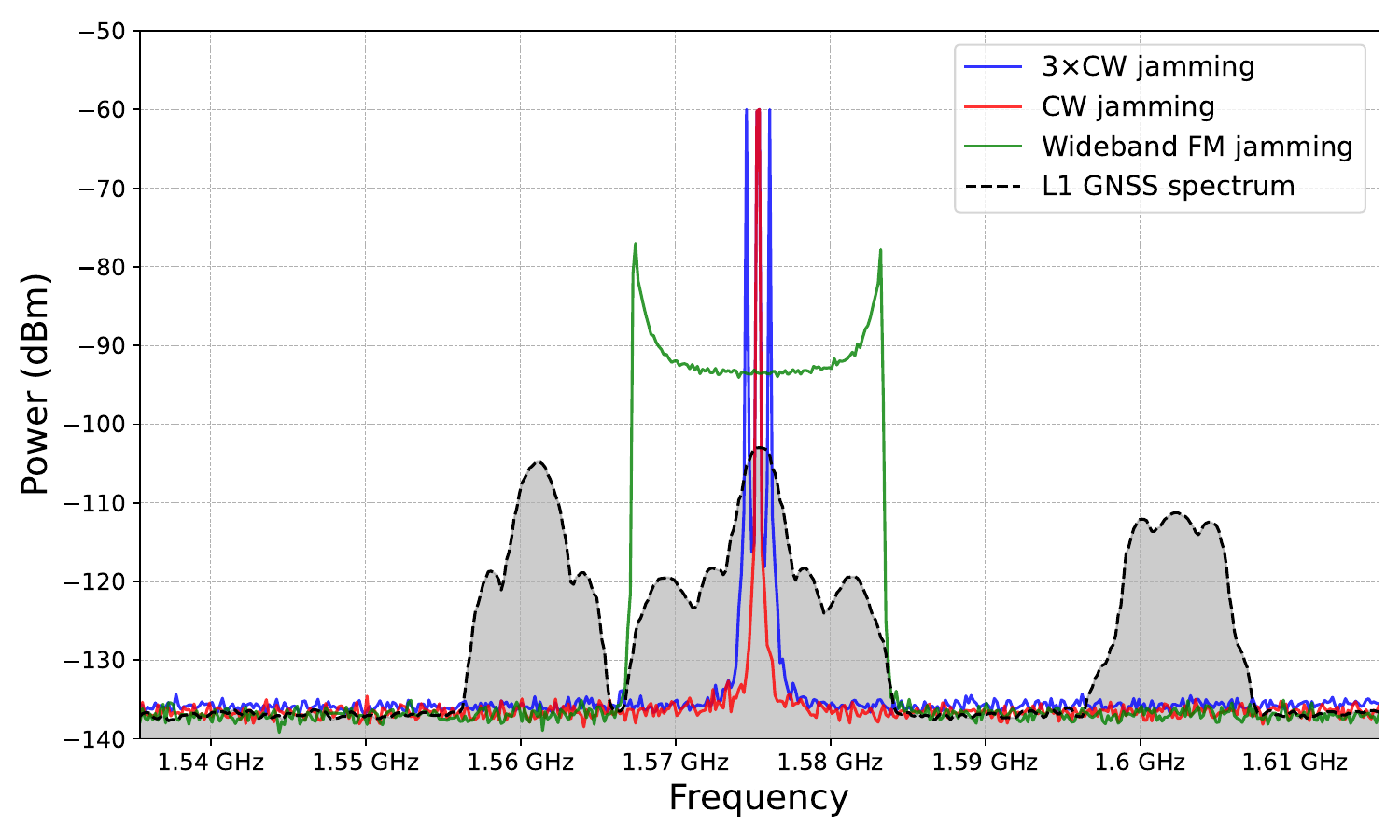}
  \caption{Three distinct jamming profiles, represented in different colors, overlaid on the L1-band GNSS spectrum (gray shaded area), illustrating the significant power disparity between the interference and the nominal signal.} 
  \label{fig:jammingspectrum}
\end{figure}

Figure~\ref{fig:jammingspectrum} shows three jamming profiles overlaid on the L1 spectrum~\cite{blatnik2025evaluating}. The dashed black line indicates the standard \gls{gnss} spectrum, while the colored traces represent different jamming profiles. Shown in red trace is a Continuous Wave signal (CW), a very narrow-band, single-tone jammer that appears as a sharp spectral spike. The blue trace is a multi-tone $3\times$CW signal, comprising three closely spaced tones that broaden the affected region and increase the likelihood of disrupting multiple satellites simultaneously. The green trace depicts wide-band FM jamming, a frequency-modulated emission spanning tens of MHz that broadly raises the noise floor, degrading all channels together. \rev{These three profiles differ in what the receiver can do against them: a single tone is attenuated by despreading and by a notch filter, three tones widen the band that a notch filter would have to cover, and a wideband modulation is suppressed by neither~\cite{kaplan2017gnss, borio2016jammers}.} 

We seek a model $\Phi$ for position error mitigation that (a) operates under CW, $3\times$CW, and wideband FM jamming, (b) remains robust across power levels, (c) generalizes \rev{to locations not seen in training \revg{in the static scenario} and to new repetitions of known routes \revg{in the kinematic scenario}, and (d) identifies the jammer type and power}.

At epoch $t$, let $S_t$ denote the set of satellites visible to the receiver, each providing features such as \gls{snr}, azimuth, and elevation, alongside the receiver's \rev{own state $\mathbf{r}_t$, which contains its} estimated position $(\text{lat}_t, \text{lon}_t)$ \rev{and, \revg{in the kinematic scenario}, its ground speed $v_t$ and heading of motion $\psi_t$}. The complete observation set is:
\begin{equation}
    O_t = \{ {\color{blue}\mathbf{r}_t}, \, o_{t,1}, \ldots, o_{t,|S_t|} \},
\end{equation}

where each $o_{t,i} \in \mathbb{R}^d$ is a $d$-dimensional feature vector containing the \gls{snr}, azimuth, and elevation corresponding to the $i$-th satellite. \rev{\revg{In the static scenario}, jamming shifts the reported position and, at higher power, also disperses it~\cite{blatnik2025evaluating}. \revg{In the kinematic scenario}, the ground speed and heading carry part of the correction (Section~\ref{sec:res:ablation}).}

Interference affects these measurements in a time-correlated manner: the \gls{snr} drops, tracking is intermittently lost, and the position drifts. Relying on a single time step $O_t$ ignores this temporal structure, so we use a short history to estimate the deviation vector $Y_t = (\Delta \text{lat}_t, \Delta \text{lon}_t)$ \rev{together with the jamming setting $c_t$, which is either no jamming or one of the three jammer types at one of the six power levels (19 classes)}:
\begin{equation}
    {\color{blue}(}\hat{Y}_t{\color{blue}, \hat{c}_t)} = \Phi(\{O_{t-k}, \ldots, O_t\})
    \label{eq:formulation}
\end{equation}
Our model relies exclusively on \rev{quantities available in the} standard NMEA \rev{output}. This minimalist feature selection emulates a worst-case scenario \rev{for the model.}

\section{Proposed Method}
\label{sec:transf}

Treating $\{O_{t-k}, \dots, O_t\}$ as a multivariate time series, as defined in Section~\ref{sec:prob:formulation}, is challenging because the cardinality $|S_t|$ varies as satellites rise and set and jamming causes additional dropouts, producing irregularly structured sequences.
\begin{figure}[th] 
  \centering
  \includegraphics[width=\linewidth]{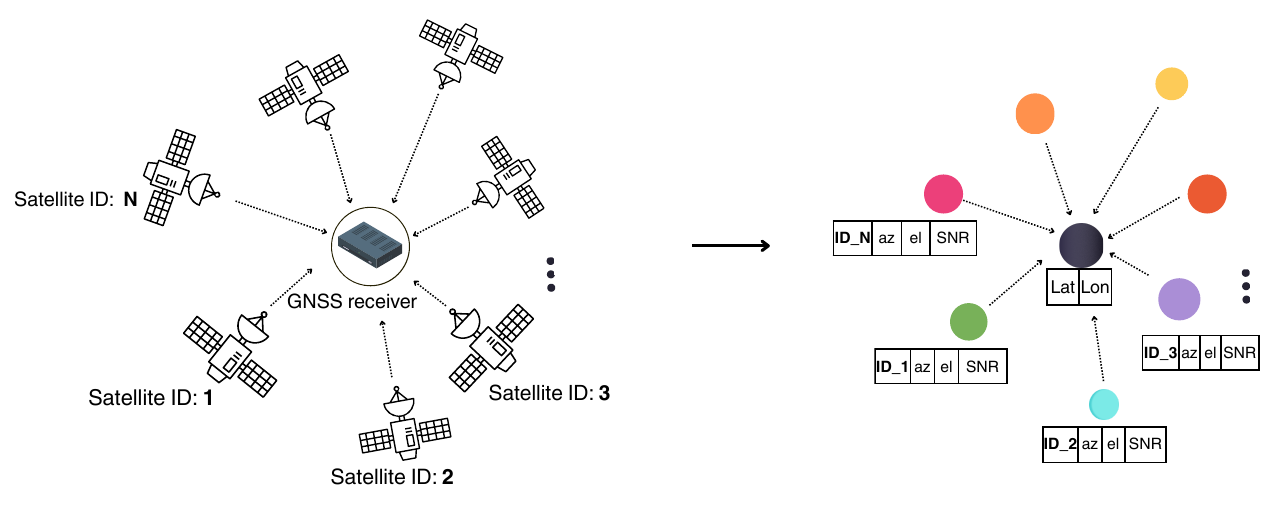}
  \caption{Transformation of the physical GNSS constellation into a heterogeneous star-graph snapshot. The receiver is modeled as the central node, while visible satellites $S_t$ are represented as leaf nodes.}
  \label{fig:satnetwork}
\end{figure}

A more natural perspective arises from the physical configuration: the receiver--satellite constellation inherently forms a star graph, with the receiver as the central node connected to visible satellites. At each epoch $t$, $O_t$ maps to a heterogeneous star-graph snapshot $G_t$ (Fig.~\ref{fig:satnetwork}). Its node set consists of a single receiver node together with $S_t \subseteq \mathcal{V}_{sat}$, the satellites visible at epoch $t$, where $\mathcal{V}_{sat}$ is the full set tracked across the window. $E_t$ contains receiver-to-satellite edges, and $X_t$ stores node features: \rev{the state $\mathbf{r}_t$} for the receiver and \gls{snr}, azimuth and elevation for each satellite.

\begin{figure}[htbp] 
  \centering
  \begin{subfigure}[t]{0.9\linewidth}
    \centering
    \includegraphics[width=\linewidth]{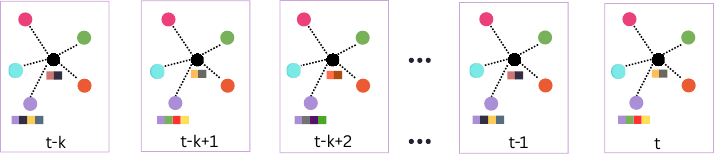}
    \caption{Normal operating conditions - stable topology and smoothly evolving features.}
    \label{fig:nojam}
  \end{subfigure}

  \vspace{0.5em}

  \begin{subfigure}[t]{0.9\linewidth}
    \centering
    \includegraphics[width=\linewidth]{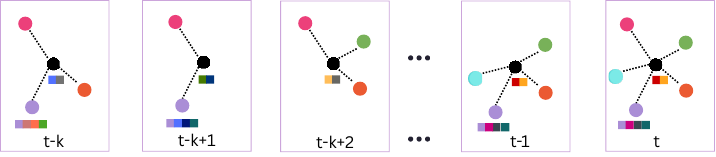}
    \caption{Jamming period - highly erratic structure characterized by sudden satellite dropouts and reacquisitions.}
    \label{fig:jamperiod}
  \end{subfigure}
  \caption{Comparison of constellation temporal graph sequences over a sliding window from $t-k$ to $t$.}
  \label{fig:temporalseq}
\end{figure}

To capture temporal evolution, we adopt the \gls{dtdg} perspective~\cite{kazemi2020representation}, representing the system as a sequence of graph snapshots $\{G_{t-k}, \dots, G_t\}$. 
Figure~\ref{fig:nojam} shows a sequence of static graph snapshots $\{G_{t-k}, \dots, G_t\}$ under normal operational conditions. In this regime, the overall network topology is relatively stable, and the node attributes, represented by the colored bars next to each node, evolve in a continuous and smooth manner.
In contrast, Figure~\ref{fig:jamperiod} illustrates a sequence of static graph snapshots $\{G_{t-k}, \dots, G_t\}$ during a jamming period. Here, the topological and feature evolution becomes abrupt and erratic. Satellite tracking status is intermittently lost, causing the edges between the receiver and satellites to disappear suddenly. 
Simultaneously, the satellite feature vectors, represented as colored bars, become volatile, often showing a rapid drop of measurements such as the \gls{snr}, and the central receiver node's features also become less stable. 

This structured graph representation allows us to frame the mitigation task from Section~\ref{sec:prob:formulation} as a \textbf{temporal dynamic graph regression problem}.

\subsection{Proposed Model}
\label{sec:model}

\begin{figure}[htbp]
 \centering
 \includegraphics[width=0.75\linewidth]{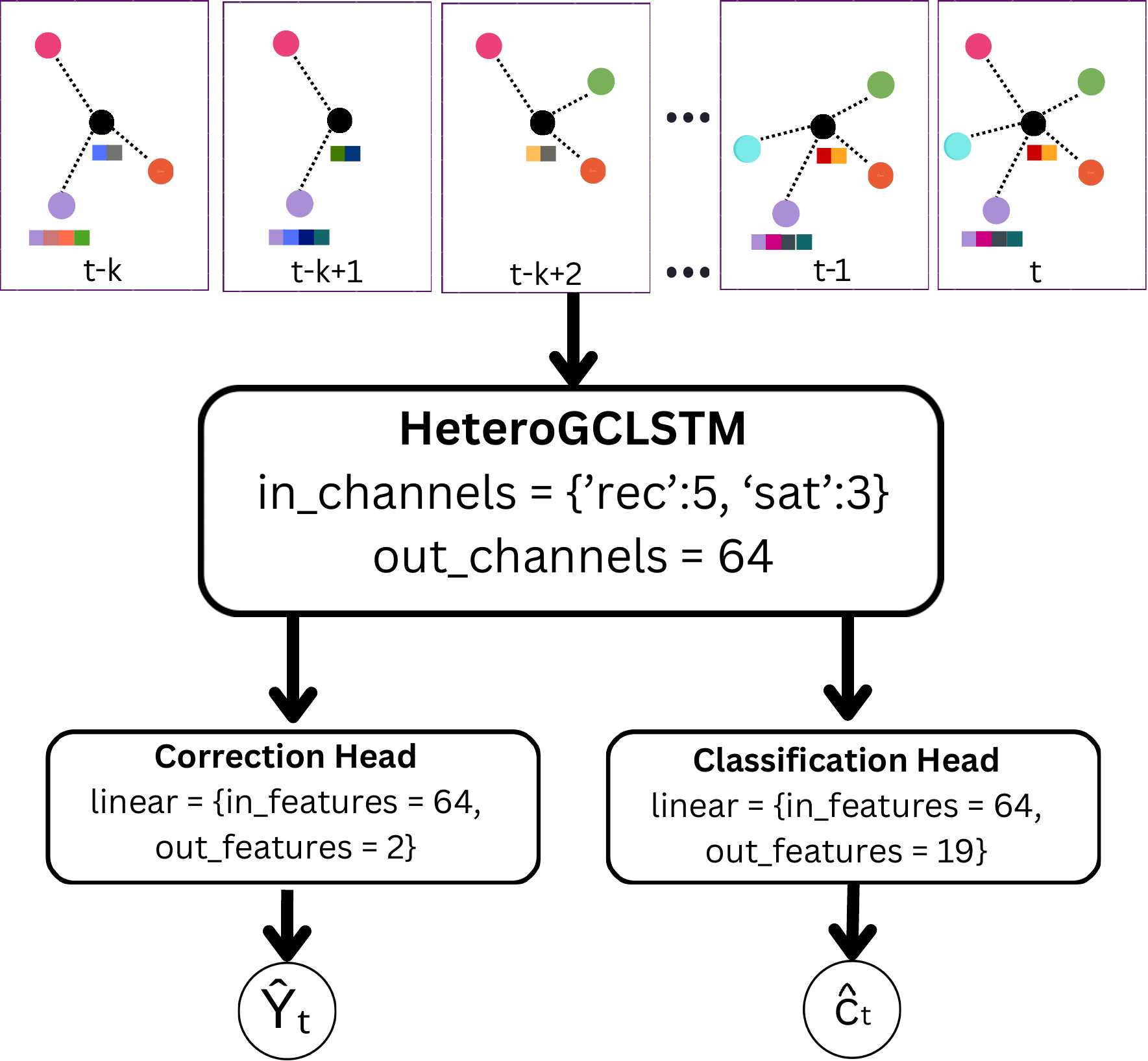}
 \caption{\revg{Architecture of \gls{jaguard}. A single HeteroGCLSTM layer processes the last $k$ graph snapshots, and the 64-dimensional state of the receiver feeds the classification  head ($\hat{c}_t$) to identify jamming and the correction head ($\hat{Y}_t$) to do PEC. Absent (hollow) satellites keep their frozen states.}}
 \label{fig:model}
\end{figure}

To develop the regression function $\Phi$ from Eq.~\ref{eq:formulation}, we design \gls{jaguard}, a spatiotemporal \gls{gnn} that captures both the geometric structure and temporal evolution of \gls{gnss} data under interference. Each input sequence consists of $k$ graph snapshots $G_t$, each containing one receiver node and a time-varying set of satellite nodes $S_t$. During jamming, satellites may temporarily disappear or reappear, causing the node set and edge structure to change between consecutive snapshots.

The architecture employs a single \gls{heterogclstm} layer~\cite{rozemberczki2021pytorch}, a choice justified by the star-graph topology. \rev{\label{rev:R1.5}We prefer this recurrent graph layer to attention-based spatio-temporal graph models for two reasons. First, the graph changes at every epoch, and a recurrent state per node follows each satellite through these changes (Eq.~\ref{eq:piecewise}). Second, the star graph has a single hop, so that attention can only reweight the satellites around one receiver.} Bidirectional edges ($\text{sat}{\to}\text{rec}$ and $\text{rec}{\to}\text{sat}$) enable the receiver to aggregate information from its one-hop neighborhood and broadcast its state back to satellites in one message-passing step. A second layer would reprocess the same one-hop messages with no additional structural information. Specifically, four HeteroConv sub-layers produce the inputs for the LSTM gates (input $\mathbf{i}$, forget $\mathbf{f}$, output $\mathbf{o}$, and cell candidate $\tilde{\mathbf{c}}$). Because there are no inter-satellite links in our receiver-centric star graph, the bipartite message passing is dynamically bounded by the active topology. For a node of type $\tau \in \{\text{rec}, \text{sat}\}$ at epoch $t$, the activation for any gate $\mathbf{g}$ explicitly fuses the current input features $\mathbf{x}^{(t)}$ with the GraphSAGE~\cite{hamilton2017inductive}-aggregated temporal hidden state $\mathbf{h}^{(t-1)}$. For the receiver node type, which aggregates the messages from the surviving satellites $S_t$, $\mathbf{g}_{\text{rec}}^{(t)}$ denotes the activation of any gate $\mathbf{g} \in {\mathbf{i}, \mathbf{f}, \mathbf{o}, \tilde{\mathbf{c}}}$ at epoch $t$, formalized as:
%For the receiver node type, which collapses messages from the surviving satellites $S_t$, this is formalized as:

\begin{equation}
\begin{split}
\mathbf{g}_{\text{rec}}^{(t)} = \sigma \bigg( &\mathbf{W}_{g, x}^{(\text{rec})} \mathbf{x}_{\text{rec}}^{(t)} + \mathbf{W}_{g, \text{rec}}^{(\text{rec})} \mathbf{h}_{\text{rec}}^{(t-1)} \\
&+ \mathbf{W}_{g, \text{sat}}^{(\text{rec})} \frac{1}{|S_t|} \sum_{s \in S_t} \mathbf{h}_s^{(t-1)} + \mathbf{b}_g^{(\text{rec})} \bigg)
\end{split}
\label{eq:gate_activation}
\end{equation}

\noindent where $\mathbf{W}_{g,x}$, $\mathbf{W}_{g,\text{rec}}$, and $\mathbf{W}_{g,\text{sat}}$ are learnable weight matrices for the input observables, the receiver's prior state, and the mean-aggregated satellite neighborhood, respectively, and $\mathbf{b}_g$ is the bias. 
For the cell candidate $\tilde{\mathbf{c}}_{\text{rec}}^{(t)}$, $\sigma$ is replaced by $\tanh$. Mean pooling over $S_t$ is permutation-invariant and handles variable constellation sizes without zero-padding. \rev{\label{rev:R1.4}It weights all satellites equally, although their contributions to the position differ with geometry and signal quality. Replacing the mean by attention-based aggregation, which weights the satellites individually, did not lower the error (Section~\ref{sec:res:ablation}).} Each active satellite $s \in S_t$ follows the same gate structure as Eq.~\ref{eq:gate_activation} but with satellite-type weights $\mathbf{W}^{(\text{sat})}$ and a single neighbor: the receiver's prior hidden state $\mathbf{h}_{\text{rec}}^{(t-1)}$ replaces the mean-pooled term, since there are no inter-satellite edges.
After computing these gates, the state updates follow standard LSTM dynamics independently for each node type. For the receiver, the updates are formalized as:
{\color{blue}\begin{equation}
\mathbf{c}_{\text{rec}}^{(t)} = \mathbf{f}_{\text{rec}}^{(t)} \odot \mathbf{c}_{\text{rec}}^{(t-1)} + \mathbf{i}_{\text{rec}}^{(t)} \odot \tilde{\mathbf{c}}_{\text{rec}}^{(t)}, \quad
\mathbf{h}_{\text{rec}}^{(t)} = \mathbf{o}_{\text{rec}}^{(t)} \odot \tanh\big(\mathbf{c}_{\text{rec}}^{(t)}\big),
\label{eq:lstm_update}
\end{equation}}
\rev{where $\odot$ denotes the element-wise product.}

%Let $\mathcal{V}_{sat}$ be the global set of all satellites tracked across the observation window, and $S_t \subseteq \mathcal{V}_{sat}$ the subset visible at epoch $t$. 
To preserve temporal context during signal outages, each satellite $s \in \mathcal{V}_{sat}$ follows a piecewise state update:

\begin{equation}
\left(\mathbf{h}_s^{(t)}, \mathbf{c}_s^{(t)}\right) =
\begin{cases}
  \begin{array}{@{}l@{}}
    \text{HGCLSTM}^{(\text{sat})}\big(\mathbf{x}_s^{(t)}, \mathbf{h}_s^{(t-1)}, \\
    \quad \mathbf{c}_s^{(t-1)}, \mathbf{h}_{\text{rec}}^{(t-1)}\big)
  \end{array} & s \in S_t, \\[14pt]
  \left(\mathbf{h}_s^{(t-1)}, \mathbf{c}_s^{(t-1)}\right) & s \notin S_t.
\end{cases}
\label{eq:piecewise}
\end{equation}

For any occluded or jammed satellite $s \notin S_t$, the hidden and cell states are frozen rather than zeroed. Upon reacquisition at epoch $t' > t$, the satellite resumes from its preserved state, restoring pre-outage temporal context without re-initialization. Because the model operates on 10-second windows, the satellite’s orbital geometric shift during any intra-window dropout is negligible ($<0.1^\circ$), making the frozen state physically valid for tracking the RF degradation profile. In kinematic scenarios, while vehicle
translation during an outage means the frozen state carries RF interference context rather than instantaneous spatial geometry, it preserves the continuous signal history required by the identification head.

\begin{figure*}[!t]
\centering
\includegraphics[width=\textwidth]{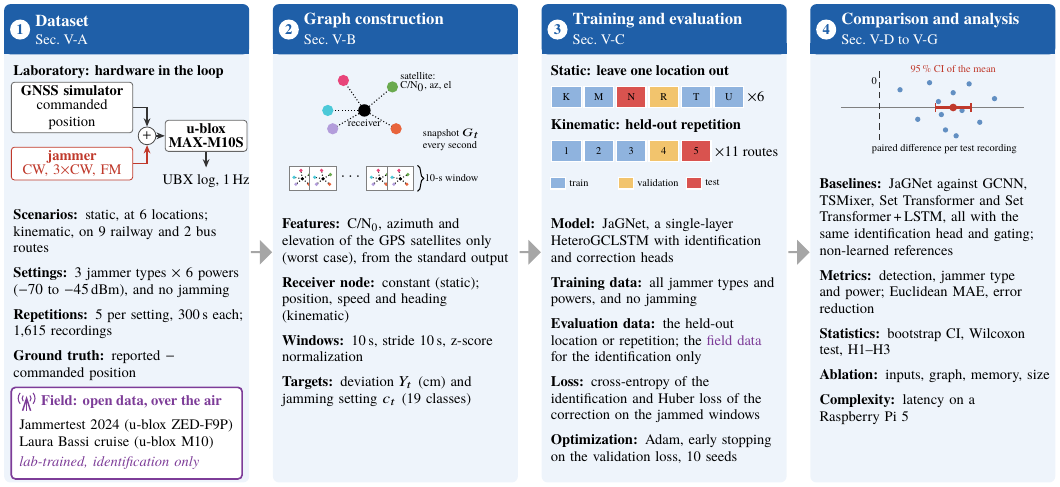}
\caption{\revg{Methodology: laboratory and open field data (1), graph construction (2), training and held-out evaluation (3), and comparison of \acrshort{jaguard} with the baselines (4).}}
\label{fig:method}
\end{figure*}

As shown in Fig.~\ref{fig:model}, the model unrolls through $k$ snapshots, mapping inputs into a \rev{64}-dimensional hidden space\rev{\label{rev:R1.6}, a size selected in the ablation study (Section~\ref{sec:res:ablation})}. The receiver's final hidden state, produced after unrolling through all $k$ snapshots, is passed to a regression head: dropout ($p=0.1$) followed by a linear layer producing the 2D output $\hat{Y}_t$. No activation function is applied, as range-restricting nonlinearities would limit the ability to represent the full range of real-valued positional deviations.

\rev{After the same dropout, the receiver's final hidden state is passed to an identification head, a linear layer that produces the scores of the 19 classes of $c_t$. Both heads are trained jointly (Section~\ref{sec:training}). At inference, the reported position is corrected by $\hat{Y}_t$ only when the most probable class is a jammed one, and otherwise it is passed on unchanged. \label{rev:R3.1}\revg{In the static scenario}, the receiver node has a constant input and carries no coordinates, so the model cannot learn the location from the reported position. \revg{In the kinematic scenario}, it carries the reported position, the ground speed and the sine and cosine of the heading of motion.}

\section{Methodology and Experimental Details}
\label{sec:expsetup}
This section describes the dataset, \rev{the graph construction, the} training procedure, and evaluation \revg{and analysis}\rev{, summarized in Fig.~\ref{fig:method}}.

\subsection{Dataset}
\label{sec:dataset}

\revg{\textit{Laboratory recordings.}} To evaluate our method, we \rev{recorded a new dataset that} comprises multivariate time series of \gls{gnss} observations collected through \rev{the} conducted testing framework \rev{of Blatnik and Batagelj~\cite{blatnik2025evaluating}, extended from a single fixed location to \revg{a static scenario with} six locations and \revg{a kinematic scenario with} eleven routes} \revg{(Fig.~\ref{fig:method}, stage~1)}. As shown in the Data column of Table~\ref{tab:related}, jamming detection studies use simulated~\cite{zhong2024tsfanet, mehr2024deep} or proprietary~\cite{spanghero2025gnss} receiver data, while existing open-data PEC works~\cite{kanhere2022improving, zhu2024enhancing, mohanty2023learning} target urban multipath rather than jamming. To our knowledge, this is the first work to evaluate jamming-induced position error correction on an open, controlled \rev{dataset that exposes one receiver to the same interference at rest and in motion and will be released upon acceptance}, enabling reproducible benchmarking. \rev{\label{rev:R1.2}Data were collected with a u-blox MAX-M10S receiver (firmware SPG~5.10), a low-cost unit widely used in fleet-tracking and transport applications~\cite{hamza2025recent}. The simulator generates the signals of all constellations, and the receiver computes a standard single-point position from its GPS, Galileo and BeiDou measurements, without augmentation, differential or carrier-phase corrections. Its output was logged in the u-blox binary protocol (UBX) at 5\,Hz and down-sampled to whole GPS seconds.}

The experimental framework employed synthesized RF interference to establish a controlled testing environment. Interference was introduced with three jamming modalities: Continuous Wave (CW), multi-tone Continuous Wave ($3\times$CW), and wideband FM, at six power levels from $-45$ to $-70$\,dBm\rev{, set at the output of the jamming generator}. \rev{\label{rev:R1.3}White-noise jamming was not included, because its effect on a spread-spectrum receiver, a reduction of the effective carrier-to-noise density ratio of all tracked signals~\cite{kaplan2017gnss}, is also produced by the wideband FM jamming.}

\rev{The simulator generates the satellite signals for a commanded receiver position and feeds them to the receiver by cable, together with the jamming signal~\cite{blatnik2025evaluating}. The satellite signals are generated with Keysight Signal Studio for GNSS (N7609C) and a Keysight N5182B signal generator at $-130$\,dBm, and the jamming signal of a second N5182B is added through a directional coupler: one tone at 1575.42\,MHz (CW), three tones at 1575.42\,MHz and $\pm$735\,kHz ($3\times$CW), and a frequency modulation around 1575.42\,MHz with 8\,MHz deviation and 1\,kHz rate (FM)~\cite{blatnik2025evaluating}.} Because the RF environment was fully synthesized, \rev{the commanded position} serves as the absolute external ground-truth\rev{: a fixed coordinate at each of the six locations \revg{of the static scenario}, Koper, Mrakovica, Nanos, Ro\v{z}nik, Triglav and ULFE, in Slovenia and Bosnia and Herzegovina, and a predefined trajectory on each of the eleven routes \revg{of the kinematic scenario}, nine railway lines and two bus lines (Table~\ref{tab:routes}), aligned with the receiver output by GPS time}. \revg{Along a route, the simulated signals carry the Doppler shift and the changing satellite visibility of the moving receiver.} The true metric deviations ($\Delta lat$, $\Delta lon$) \rev{are the differences between the reported and the commanded position at each epoch, in centimeters. The conducted setup isolates the effect of the jamming from multipath, obstruction and antenna effects (Section~\ref{sec:limitations}). \label{rev:R2.6-epoch}All recordings were simulated at the same date and time.}
\begin{figure}[h]
\centering
\includegraphics[width=0.9\linewidth]{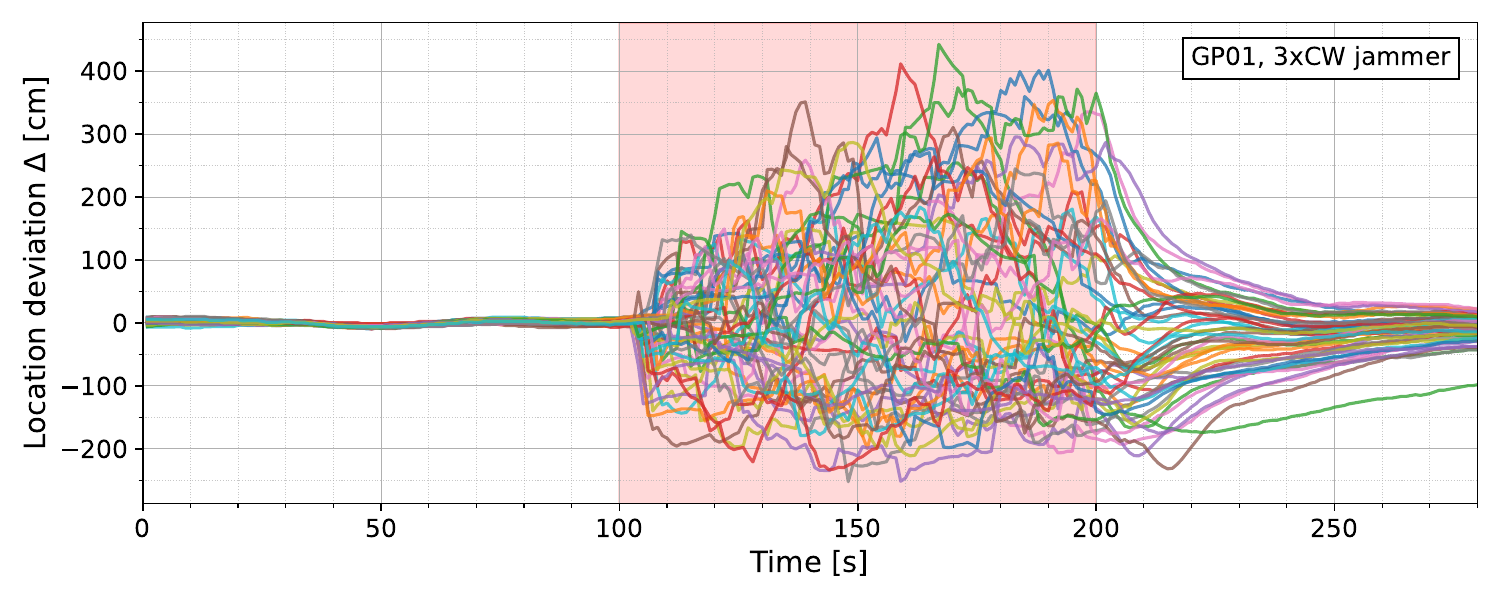}
\caption{Positional deviation of \rev{a} GNSS receiver across 50 repeated measurements under $3\times$CW jamming \rev{from 100 to 200\,s (shaded), adapted from~\cite{blatnik2025evaluating}}.}
\label{fig:variability}
\end{figure}

To characterize receiver stability, every unique combination of \rev{location or route}, jammer, and power level was repeated \rev{five} times\rev{, together with five recordings without jamming, which gives 95 recordings per location or route and 1,615 in total}. Each repetition followed a strict temporal protocol: \rev{after a restart of the receiver and a warm-up until a fix was available, the jammer was switched on, and the receiver output was recorded for the following 300\,s. On a route, every repetition covers the same stretch of the trajectory}. Data were sampled at \rev{1\,Hz}, producing a time-ordered multivariate sequence that reflects evolving satellite visibility and signal quality~\cite{blatnik2025evaluating}. While in theory these repetitions of a deterministic process should yield identical results, in practice, the receiver's internal logic and real-time processing introduce behavioral variations in each run.

The inherent unpredictability of the \gls{gnss} receiver’s response to a disturbance is illustrated in Figure~\ref{fig:variability}. It presents 50 repeated measurements\rev{~\cite{blatnik2025evaluating}} of \rev{a} receiver’s positional deviation in response to a jamming signal, present at times denoted as the shaded red area. Each colored line represents an individual measurement. \rev{Before the jamming, the repetitions coincide.} Despite constant receiver location and satellite geometry, the resulting positional drift \rev{under jamming differs from one repetition to the next}. \rev{This variation arises in the proprietary firmware of the receiver, whose response depends on which of its mitigation methods engage and on the processing step it is in when the jamming starts~\cite{blatnik2025evaluating}.} \rev{\label{rev:offset}Our recordings show the same behavior: under $3\times$CW jamming at $-50$\,dBm at the static location Nanos, the MAE of the five repetitions ranges from 137.7 to 216.9\,cm.}

\rev{\label{rev:field-data}\textit{Field recordings.} The \revg{detection} needs no reference position, so it can be tested over the air and with other receivers on two open datasets. The Jammertest 2024 dataset~\cite{sayyaf2025jammertest} contains the output of a u-blox ZED-F9P at the Jammertest event~\cite{gerrard2026jammertest}: at rest, under a jammer whose power is ramped up, and in motion, in five scenarios with low-power commercial jammers in vehicles. The attack log of the event gives the jamming periods. We use the 1,100 windows of 10\,s with a fix from six scenarios and leave out a seventh, whose degradation outlasts its logged jamming period. \revg{From the Laura Bassi dataset~\cite{bertalanic2025voyage}, the u-blox M10 output of an icebreaker cruise to Antarctica, we use 180 days with 153 interference events, labeled with an additional GNSS receiver on board, and a random sample of 4,539 minutes labeled clean.} The models for these tests are trained on the static laboratory recordings and use the GPS satellites only (C/N$_0$, azimuth, elevation), without receiver inputs.}

\subsection{\revsec{Graph Construction}}
\label{sec:features}

\rev{The features are decoded from the UBX-NAV-PVT and UBX-NAV-SAT messages and are equally available in the standard NMEA sentences GGA, RMC, VTG and GSV~\cite{ublox_m10_ifd} \revg{(Fig.~\ref{fig:method}, stage~2)}. The \gls{snr} is the carrier-to-noise density ratio C/N$_0$ that the receiver reports in dB-Hz. \label{rev:R2.4}The satellite features are restricted to GPS as the worst case: under $3\times$CW jamming and FM jamming from $-45$ to $-60$\,dBm, at most two GPS satellites remain in the solution on average, and the receiver computes the position mainly from BeiDou satellites. \gls{jaguard} thus corrects the position of a multi-constellation receiver from its GPS satellites alone. Section~\ref{sec:res:gec} repeats the evaluation with the satellites of all constellations.}

To convert these raw time-series sequences into inputs for spatio-temporal learning, each timestamp is transformed into a structured graph snapshot as detailed in Section~\ref{sec:transf}. This snapshot-based formulation naturally encodes both the heterogeneity of the interaction space, with different node types and distinct feature sets, and its dynamics, including time-varying topology and signal attributes, within the satellite-receiver interaction space. Input sequences were generated using a sliding window with size and stride of 10, selected via an ablation study. All features and targets were Z-score normalized prior to training. Final metrics are reported after inverse transformation to physical centimeters.

\subsection{\revsec{Training and Evaluation}}
\label{sec:training}

\rev{\textbf{\revg{Static scenario.}} Each of the six locations is held out for testing in turn \revg{(Fig.~\ref{fig:method}, stage~3)}. The next location in alphabetical order is used for validation and the remaining four for training (leave-one-location-out). \textbf{\revg{Kinematic scenario.}} Trains and buses travel the same route repeatedly. Following the fingerprinting scenario of~\cite{weng2024prnet}, the model of each route is therefore trained on repetitions 1 to 3 of every setting, validated on repetition 4 and tested on repetition 5, a complete repetition not seen in training.} To emulate operational conditions where the type and strength of interference vary unpredictably, \rev{every model is trained on all jammer types and power levels together with the recordings without jamming, which are balanced by design (five recordings per setting)}.

We used the Adam optimizer with learning rate $10^{-3}$ and weight decay $10^{-2}$ for L2 regularization. The loss function is Smooth L1 (Huber) Loss with $\beta=10^{-2}$, which grows linearly for large errors, providing stability against the sudden deviations characteristic of jammed signals. \rev{The two-head models minimize the sum of the cross-entropy of the identification over all windows and the Huber loss of the correction over the jammed windows, whose targets are normalized with the statistics of the jammed training recordings. They are trained with mini-batches of 16 windows for at most 60 epochs, with early stopping (patience of six epochs) on the same loss over the validation recordings.}

Early stopping saves the best validation checkpoint. Each model was trained with ten random seeds, where the results report averages \rev{over the seeds and, where indicated, the \gls{std} over the held-out locations}.

\subsection{Comparison to the Baseline Models}
\label{sec:baseline}

We compare \gls{jaguard} against one state-of-the-art (SotA) model and three baselines \revg{(Fig.~\ref{fig:method}, stage~4)}. All models predict 2D positional deviation from the same NMEA features, are trained with Adam, Huber loss, and Early Stopping across 10 seeds. \rev{\label{rev:R3.3}Every baseline is trained with the same identification head and gating as \gls{jaguard}, so that all models solve the same task and the comparison isolates the architecture. The baselines with a receiver representation receive the same receiver input as \gls{jaguard}, and all models are trained on the same data, splits and targets. Their sizes range from 5,111 to 90,453 parameters, against 69,205 for \gls{jaguard} (Table~\ref{tab:computation}).} The purely spatial SotA \gls{gcnn} processes only the instantaneous geometry at time $t$, while all other models use a 10-second window, isolating the value of temporal context.

The SotA reference is a \gls{gcnn} adapted from Mohanty and Gao~\cite{mohanty2023learning}, the current graph-based PEC state-of-the-art. It uses two \gls{gin} layers on an inter-satellite graph, without an explicit receiver node, and applies global mean pooling for graph-level correction. \rev{The $\epsilon$ parameter of its \gls{gin} layers is learned, because with $\epsilon=0$ on the fully connected satellite graph, every node would receive the same aggregated input.} We adapt it to our NMEA feature set while retaining the original inter-satellite mesh topology. This comparison\rev{,} therefore, tests the topology choice (inter-satellite mesh vs. receiver-centric star) and the temporal modeling (single-epoch vs. sequential), rather than claiming to reproduce the original system, which was designed for multipath with different input features. \rev{It thereby shows the combined effect of memoryless processing and of a topology without a receiver node. The ablation study separates the two (Section~\ref{sec:res:ablation}).}

TSMixer~\cite{chen2023tsmixer} is a fully-connected \rev{multilayer-perceptron} model that serves as a strong representative for time series regression tasks. It operates directly on the multivariate time series representation (\gls{snr}, azimuth, elevation\rev{, and the receiver state $\mathbf{r}_t$}) of our \gls{gnss} data. Because TSMixer is strictly an MLP-based architecture that requires fixed-size input tensors, we accommodate the dynamic number of visible satellites per epoch by zero-padding the feature vectors up to a predefined maximum constellation size\rev{, with one slot and a presence flag per GPS satellite number}.

The Set Transformer~\cite{lee2019set} processes the 10-second window as an unordered set, deliberately discarding chronological order. Stacked Set Attention Blocks (SAB) capture satellite interactions, with distinct projections for satellite (3D) and receiver \rev{($\mathbf{r}_t$)} features into a shared 64-dimensional space. Our adaptation uses the receiver state as the PMA query instead of learnable seeds, conditioning aggregation on the receiver’s context.

The Temporal Set Transformer (TST)\rev{, denoted Set Transformer + LSTM in the tables,} extends the Set Transformer by reintroducing chronological structure through a hybrid spatial-recurrent architecture. Each one-second snapshot in the 10-second window is independently encoded by the Set Transformer (SAB + PMA), conditioned on the receiver's instantaneous state, producing a latent vector. The resulting sequence of ten embeddings is passed to an LSTM, whose final hidden state yields the prediction. During a total \gls{gnss} outage, a receiver-only fallback maintains temporal continuity across observation gaps.

\rev{In addition, we report three references that are not learned: the uncorrected position, the subtraction of one mean offset of all jammed training recordings, and the subtraction of the mean offset of the training recordings of the true jamming setting, which serves as an oracle.}

\subsection{\revsec{Metrics and Statistical Analysis}}
\label{sec:metrics}

\rev{\label{rev:ident-metrics}Both outputs are evaluated on the 10-s test windows at their last epoch, including all jammed windows \revg{(Fig.~\ref{fig:method}, stage~4)}. The detection rate is the share of jammed windows identified as jammed, and the \revg{false-positive rate} the share of unjammed windows identified as jammed. The \revg{identification} is evaluated on the jammed windows by three accuracies: the jammer type is correct (type accuracy), the type is correct and the power lies within 5\,dB of the true power, i.e., at the true or an adjacent level (type and power within 5\,dB), and the type and power are both correct (exact-setting accuracy). As a reference, the receiver's own interference monitor reports a jamming state in the UBX-MON-RF message~\cite{ublox_m10_ifd}, and a window counts as flagged when any of its messages reports the state \emph{warning} or \emph{critical}. The field recordings have no jammer settings, so only the detection is evaluated there. On Jammertest 2024, the 10-s windows are grouped by the drop of the mean GPS C/N$_0$ from its mean before the jamming of the same scenario. \revg{Windows that end more than 60\,s before the logged start of the jamming are clean, so the share flagged among them is the \revg{false-positive rate}, and windows that end within 60\,s of the logged start or end are left out.} On the cruise, a minute counts as flagged when any of its windows is flagged, an interference event counts as detected when any of its minutes is flagged, the events are grouped by their largest C/N$_0$ drop, and the \revg{false-positive rate} is the share of clean minutes flagged.} We evaluate \rev{the correction} using Mean Absolute Error (MAE) for latitude and longitude deviations:
$MAE_{lat} = \frac{1}{M}\sum_{i=1}^{M} |\Delta lat_i - \Delta\widehat{lat}_i|$
and $MAE_{lon} = \frac{1}{M}\sum_{i=1}^{M} |\Delta lon_i - \Delta\widehat{lon}_i|$,
where $M$ is the number of evaluation samples. 
We also compute the Euclidean MAE as a unified 2D error measure:
$MAE_{euc} = \frac{1}{M}\sum_{i=1}^{M} \sqrt{(\Delta lat_{i}-\Delta\widehat{lat}_{i})^2 + (\Delta lon_{i}-\Delta\widehat{lon}_{i})^2}$.
\noindent All \rev{errors} are reported in centimeters.

\rev{The error reduction is $R = 1 - \mathrm{MAE}_{euc}/\mathrm{MAE}_{euc}^{\mathrm{raw}}$, where $\mathrm{MAE}^{\mathrm{raw}}_{euc}$ is the MAE of the uncorrected position over the same windows. For means over locations or routes, the ratio is formed after averaging.}

\rev{\label{rev:R3.5}\noindent\textbf{Statistical analysis.} To separate the variability of the measurements from that of the training, every comparison is made on complete test recordings: the MAE of each recording is averaged over the seeds, and the difference between two models is computed per recording. We report the mean difference with a 95\% percentile bootstrap confidence interval (CI) from 10,000 resamples~\cite{efron1993bootstrap} and the $p$-value of a two-sided Wilcoxon signed-rank test~\cite{wilcoxon1945individual}. Because the recordings of one location or route are not independent, we also report at how many locations or routes the difference has the same sign. \revg{Within each jamming setting at rest and each route, the same paired bootstrap marks the models whose MAE is not significantly above the lowest.}}

\subsection{Ablation Study}
\label{sec:ablation}

To evaluate key hyperparameters and justify our architectural choices (Section~\ref{sec:model} \revg{and Fig.~\ref{fig:method}, stage~4}), we \rev{trained variants of the two-head model that change one element at a time, with all other settings fixed, on the static \revg{scenario} (leave-one-location-out, all settings) and on the bus route FE--UKC (within the route).}

\rev{\label{rev:R2.2}\label{rev:R3.4}The variants change the inputs, the graph, the memory or the size of the model. The input variants remove the satellite features one at a time and together, add the receiver coordinates at rest, and, on the route, remove the receiver inputs in part or in full. The graph variants remove the edges ($W=10$) or the history ($W=1$), replace the star by a satellite mesh without a receiver node over one or ten epochs, as in~\cite{mohanty2023learning}, keep only the satellite-to-receiver edges, add inter-satellite edges, or replace the mean aggregation by attention (GATv2~\cite{brody2022attentive}). The memory variants replace the frozen per-satellite memory by a reset at each outage or by one state shared by all satellites. The size variants change the window to 5, 20 or 30 epochs and the hidden size from 16 to 256. Each variant is compared with the proposed model per test recording, in the MAE and in the identification accuracy, on the windows that end at least 30\,s after jammer switch-on, so that all window lengths are evaluated on the same windows.} Since the star topology limits the useful depth to a single layer (Section~\ref{sec:model}), we focused on width rather than depth.

\subsection{Computational Complexity Evaluation}
\label{sec:complexity}

To assess operational feasibility, we evaluated parameter counts, \rev{the floating-point operations} and inference latency across all models \rev{in their two-head versions} \revg{(Fig.~\ref{fig:method}, stage~4)}. Latency was measured as the forward pass execution time over one 10-snapshot window, averaged across 1000 \rev{windows of the dataset after 50 warm-up windows}, with \rev{the} standard deviation \rev{and the 99th percentile} reported. \rev{\label{rev:R3.6}Benchmarking was conducted on a Raspberry Pi~5 (four Arm Cortex-A76 cores, PyTorch on the CPU, batch size 1), a single-board computer representative of embedded host platforms~\cite{chen2019edge}.}

\section{Results}
\label{sec:results}

In this section, we evaluate the performance of our proposed approach from Section~\ref{sec:transf} to solve the problem formulated in Section~\ref{sec:prob:formulation}. The details of the experimental setup used to obtain the results are described in Section~\ref{sec:expsetup}. \rev{We first report the detection and \revg{identification} of the jamming and then the correction of the position\revg{, at held-out static locations and along known kinematic routes (Sections~\ref{sec:res:overall} and~\ref{sec:res:routes}), along a new route and for jammers left out of training (Sections~\ref{sec:res:deploy} and~\ref{sec:res:leftout}), and with the satellites of all constellations (Section~\ref{sec:res:gec}), followed by the ablation and the computational cost (Sections~\ref{sec:res:ablation} and~\ref{sec:res:computation})}.}

\subsection{\revsec{Detection and Identification of the Interference}}
\label{sec:res:ident}

\begin{table}[!ht]
\color{blue}
\centering
\setlength{\tabcolsep}{2.5pt}
\caption{\revg{Identification in the laboratory (\%, pooled over ten seeds). FM: detected under FM jamming. 5\,dB: type correct, power within 5\,dB. Exact: type and power correct. Bold, over all powers except FM: highest of the column and the values not significantly below it (paired 95\% bootstrap CI, Section~\ref{sec:metrics}). $\uparrow$/$\downarrow$: higher/lower is better.}}
\resizebox{\columnwidth}{!}{%
\begin{tabular}{l ccccc ccccc}
\toprule
 & \multicolumn{5}{c}{\textbf{Static}} & \multicolumn{5}{c}{\textbf{\revg{Kinematic}}} \\
\cmidrule(lr){2-6}\cmidrule(lr){7-11}
 & \multicolumn{5}{c}{\revg{Jammed windows (\%)\,$\uparrow$}} & \multicolumn{5}{c}{\revg{Jammed windows (\%)\,$\uparrow$}} \\
\cmidrule(lr){2-6}\cmidrule(lr){7-11}
\textbf{Method} & Det. & \revg{FM} & Type & \revg{5\,dB} & \revg{Exact} & Det. & \revg{FM} & Type & \revg{5\,dB} & \revg{Exact} \\
\midrule
\multicolumn{11}{l}{\revg{\textit{\gls{jaguard}, by jammer power}}} \\
\revg{$-45$\,dBm} & 100.0 & 100.0 & 98.8 & 76.8 & 64.0 & 100.0 & 100.0 & 99.5 & 71.0 & 56.6 \\
\revg{$-50$\,dBm} & 100.0 & 100.0 & 96.6 & 83.3 & 55.1 & 100.0 & 100.0 & 97.2 & 74.5 & 48.2 \\
\revg{$-55$\,dBm} & 100.0 & 100.0 & 96.0 & 85.7 & 61.6 & 100.0 & 100.0 & 92.9 & 85.0 & 47.2 \\
\revg{$-60$\,dBm} & 100.0 & 100.0 & 99.0 & 81.5 & 61.8 & 100.0 & 100.0 & 99.0 & 78.5 & 67.6 \\
\revg{$-65$\,dBm} & 99.8 & 100.0 & 95.6 & 93.3 & 63.0 & 99.9 & 100.0 & 92.9 & 90.6 & 69.0 \\
\revg{$-70$\,dBm} & 99.5 & 100.0 & 93.0 & 89.0 & 76.6 & 98.8 & 100.0 & 89.0 & 86.1 & 74.8 \\
\midrule
\multicolumn{11}{l}{\revg{\textit{All powers}}} \\
\gls{jaguard} & \textbf{99.9} & 100.0 & 96.6 & 84.7 & 63.5 & \textbf{99.8} & 100.0 & 95.2 & 80.6 & \textbf{60.1} \\
GCNN & 99.4 & 100.0 & 88.1 & 70.0 & 46.5 & 99.6 & 100.0 & 88.5 & 68.1 & 46.1 \\
TSMixer & 99.8 & 100.0 & \textbf{97.2} & 84.8 & 62.5 & \textbf{99.8} & 100.0 & \textbf{97.5} & \textbf{82.9} & \textbf{60.6} \\
Set Transformer & 99.8 & 99.9 & 94.8 & 84.9 & 64.0 & \textbf{99.8} & 100.0 & 93.6 & 80.5 & \textbf{61.6} \\
Set Tr. + LSTM & \textbf{99.9} & 100.0 & \textbf{97.9} & \textbf{86.4} & \textbf{65.7} & \textbf{99.8} & 100.0 & 95.7 & \textbf{81.8} & \textbf{61.0} \\
Receiver monitor & 66.2 & 6.9 & -- & -- & -- & 64.7 & 3.7 & -- & -- & -- \\
\bottomrule
\end{tabular}}
\label{tab:ident}
\end{table}

\rev{\label{rev:R2.3}Table~\ref{tab:ident} compares the identification of \gls{jaguard} with the receiver's own interference monitor. \gls{jaguard} detects 99.9\% of the jammed windows at held-out locations without \revg{false positives} and 99.8\% on the routes with 1.0\% \revg{false positives}, and identifies the jammer type in 96.6 and 95.2\% of the windows. The exact setting, type and power, is identified in 63.5 and 60.1\% of the windows, and the type with a power within one level (5\,dB) in 84.7 and 80.6\%. \revg{Per jammer power, \gls{jaguard} detects at least 99.5 and 98.8\% of the jammed windows and identifies the jammer type in at least 93.0 and 89.0\%. The power is identified best at the weakest powers, $-65$ and $-70$\,dBm, where the type with a power within 5\,dB is identified in 86.1 to 93.3\% of the windows, against 71.0 to 85.7\% at $-45$ to $-60$\,dBm.} The receiver's monitor flags every window under CW and $3\times$CW jamming, but only 6.9\% (static) and 3.7\% (\revg{kinematic}) of the windows under FM jamming, and on the routes it \revg{raises 16.3\% false positives}. All baselines detect 99.4 to 99.9\% of the jammed windows\revg{, with 0.0 to 3.3\% false positives}. \revg{The temporal baselines identify the jammer type in 93.6 to 97.9\% of them, TSMixer and the Set Transformer + LSTM slightly more often than \gls{jaguard}, and the memoryless GCNN in 88.1 and 88.5\%. In the laboratory, the identification is therefore a property of the two-head formulation more than of the architecture, which matters instead for} the detection of moderate interference over the air (Table~\ref{tab:field}) and \revg{for} the correction (Section~\ref{sec:res:routes}).}

\begin{table}[!ht]
\color{blue}
\centering
\setlength{\tabcolsep}{2.5pt}
\caption{\revg{Detection over the air by models trained in the laboratory only (\%, mean over 5 seeds). FP: false-positive rate on the clean Jammertest windows and cruise minutes. Detected: detection rate by the drop of the mean GPS C/N$_0$ (dB). Gray: more than 5\% FP, so the detection rates are inflated and not compared. Bold: highest detection rate outside gray. --: not available.}}
\footnotesize
\begin{tabular}{l cccc ccc}
\toprule
 & \multicolumn{4}{c}{\textbf{Jammertest 2024}} & \multicolumn{3}{c}{\textbf{Laura Bassi cruise}} \\
\cmidrule(lr){2-5}\cmidrule(lr){6-8}
 & & \multicolumn{3}{c}{\revg{Detected\,$\uparrow$}} & & \multicolumn{2}{c}{\revg{Detected\,$\uparrow$}} \\
\cmidrule(lr){3-5}\cmidrule(lr){7-8}
\textbf{Method} & \revg{FP\,$\downarrow$} & 3--6 & 6--10 & $\geq$10 & \revg{FP\,$\downarrow$} & 3--6 & $>$6 \\
\revg{$n$} & \revg{64} & \revg{31} & \revg{41} & \revg{137} & \revg{4,539} & \revg{125} & \revg{28} \\
\midrule
\gls{jaguard} & 0.0 & \textbf{23.2} & \textbf{65.4} & 97.7 & 1.5 & \textbf{25.0} & \textbf{100.0} \\
GCNN & 0.0 & 3.2 & 42.0 & \textbf{100.0} & 0.5 & 21.4 & \textbf{100.0} \\
TSMixer & \textcolor{gray}{60.6} & \textcolor{gray}{62.6} & \textcolor{gray}{66.8} & \textcolor{gray}{94.2} & \textcolor{gray}{40.5} & \textcolor{gray}{39.2} & \textcolor{gray}{98.6} \\
Set Transformer & \textcolor{gray}{8.8} & \textcolor{gray}{18.7} & \textcolor{gray}{61.5} & \textcolor{gray}{98.8} & \textcolor{gray}{23.8} & \textcolor{gray}{44.3} & \textcolor{gray}{100.0} \\
Set Tr. + LSTM & 0.3 & 16.8 & 57.1 & 98.4 & \textcolor{gray}{10.0} & \textcolor{gray}{41.8} & \textcolor{gray}{100.0} \\
\midrule
C/N$_0$ threshold & 0.0 & 12.9 & 34.1 & 98.5 & 0.0 & 15.2 & \textbf{100.0} \\
\midrule
Receiver monitor & 0.0 & 0.0 & 0.0 & 0.0 & -- & -- & -- \\
\bottomrule
\end{tabular}
\label{tab:field}
\end{table}

\rev{\label{rev:field}Table~\ref{tab:field} reports the \revg{detection} on the field recordings. Trained in the laboratory only, \gls{jaguard} flags none of the 64 Jammertest windows before jamming and detects 97.7\% of the windows in which the mean GPS C/N$_0$ drops by 10\,dB or more, 65.4\% of those with a drop of 6 to 10\,dB and 23.2\% of those with 3 to 6\,dB. The receiver's own monitor flags none of these windows. On the cruise, \gls{jaguard} detects all 28 interference events with a C/N$_0$ drop above 6\,dB and 25.0\% of the 125 weaker ones, and it flags 1.5\% of the clean minutes. \revg{Clean signals over the air are stronger and more varied than in the laboratory (GPS C/N$_0$ of 41.6\,$\pm$\,6.0\,dB-Hz on the clean cruise minutes, against 38.0\,$\pm$\,4.2\,dB-Hz in the unjammed laboratory recordings), and the baselines do not all carry their notion of a clean signal into the field. TSMixer and the Set Transformer flag 60.6 and 8.8\% of the clean Jammertest windows, and TSMixer, the Set Transformer and the Set Transformer + LSTM flag 40.5, 23.8 and 10.0\% of the clean cruise minutes. Their higher detection of the weaker cruise events (39.2 to 44.3\%) and of weak Jammertest interference reflects these false positives: TSMixer, for example, flags the windows during weak jamming no more often than the clean ones (62.6 vs.\ 60.6\%). Among the learned models, only \gls{jaguard} and the GCNN raise no \revg{false positives} at Jammertest and at most 1.5\% on the cruise, and of the two, \gls{jaguard} detects more of the windows with a drop of 6 to 10\,dB (65.4 vs.\ 42.0\%) and of 3 to 6\,dB (23.2 vs.\ 3.2\%).} A threshold on the mean GPS C/N$_0$ of a window, set to 1\% \revg{false positives} on the unjammed laboratory training windows (34.81\,dB-Hz), detects over the air 98.5\% of the Jammertest windows with a drop of 10\,dB or more and all 28 strong cruise events, without \revg{false positives}. It detects, however, only 34.1\% of the windows with a drop of 6 to 10\,dB and 15.2\% of the weaker cruise events, against 65.4 and 25.0\% for \gls{jaguard}.}

\subsection{\revgsec{Correction at Static Held-Out Locations}}
\label{sec:res:overall}

\begin{table*}[!ht]
\color{blue}
\centering
\setlength{\tabcolsep}{4pt}
\renewcommand{\arraystretch}{1.15}
\caption{Euclidean MAE (cm) at the held-out locations per jammer type and power: mean over the six locations and ten seeds $\pm$ standard deviation over the locations. \revg{Bold: lowest MAE of the setting and the MAEs not significantly above it (paired 95\% bootstrap CI, Section~\ref{sec:metrics}).}}
\resizebox{0.95\textwidth}{!}{
\small
\begin{tabular}{ll c@{$\pm$}c c@{$\pm$}c c@{$\pm$}c c@{$\pm$}c c@{$\pm$}c c@{$\pm$}c}
\toprule
\textbf{Jamming} & \textbf{Method} \revg{(MAE\,$\downarrow$)} & \multicolumn{2}{c}{\textbf{$-$45 dBm}} & \multicolumn{2}{c}{\textbf{$-$50 dBm}} & \multicolumn{2}{c}{\textbf{$-$55 dBm}} & \multicolumn{2}{c}{\textbf{$-$60 dBm}} & \multicolumn{2}{c}{\textbf{$-$65 dBm}} & \multicolumn{2}{c}{\textbf{$-$70 dBm}} \\
\midrule
\multirow{7}{*}{CW}
& Proposed - \gls{jaguard} & \textbf{118.3} & 34.0 & \textbf{72.1} & 15.3 & 45.6 & 7.1 & \textbf{30.8} & 7.0 & \textbf{24.8} & 4.4 & \textbf{37.3} & 3.8 \\
& SotA - GCNN~\cite{mohanty2023learning} & \textbf{119.3} & 31.7 & 94.4 & 13.5 & 50.3 & 7.2 & 32.5 & 6.3 & 33.0 & 8.6 & 41.5 & 4.4 \\
& Baseline - TSMixer & \textbf{121.9} & 29.6 & 87.6 & 15.9 & 48.9 & 5.0 & 33.4 & 7.2 & 28.9 & 4.1 & \textbf{37.1} & 3.6 \\
& Baseline - Set Transformer & \textbf{118.8} & 35.1 & 74.8 & 14.6 & 44.8 & 7.1 & 31.6 & 7.0 & 25.6 & 4.8 & 38.9 & 3.3 \\
& Baseline - Set Transformer + LSTM & \textbf{119.0} & 36.0 & \textbf{69.4} & 15.0 & \textbf{43.7} & 7.8 & \textbf{30.9} & 7.0 & \textbf{24.7} & 5.1 & \textbf{37.3} & 4.3 \\
\cline{2-14}
& \textbf{Unmitigated Error (cm)} & 215.3 & 42.3 & 229.8 & 15.9 & 124.9 & 6.0 & 114.3 & 10.8 & 112.2 & 5.6 & 122.9 & 7.3 \\
& \textbf{Error Reduction (\gls{jaguard}) (\%)}\revg{\,$\uparrow$} & \multicolumn{2}{c}{45.0} & \multicolumn{2}{c}{68.6} & \multicolumn{2}{c}{63.5} & \multicolumn{2}{c}{73.1} & \multicolumn{2}{c}{77.9} & \multicolumn{2}{c}{69.7} \\
\midrule
\multirow{7}{*}{3$\times$CW}
& Proposed - \gls{jaguard} & \textbf{141.0} & 34.2 & \textbf{88.4} & 17.5 & \textbf{49.4} & 24.2 & \textbf{38.6} & 17.8 & \textbf{33.7} & 12.4 & \textbf{35.4} & 13.9 \\
& SotA - GCNN~\cite{mohanty2023learning} & 141.9 & 33.5 & \textbf{88.6} & 17.7 & \textbf{49.2} & 24.4 & \textbf{38.8} & 17.8 & \textbf{33.6} & 13.3 & \textbf{35.4} & 14.8 \\
& Baseline - TSMixer & \textbf{139.8} & 33.3 & \textbf{88.8} & 17.9 & \textbf{49.7} & 25.1 & 40.2 & 17.4 & 34.4 & 12.4 & 36.7 & 12.4 \\
& Baseline - Set Transformer & \textbf{140.8} & 33.6 & \textbf{88.3} & 17.6 & \textbf{49.4} & 23.5 & \textbf{38.4} & 17.2 & \textbf{33.3} & 12.0 & \textbf{35.1} & 12.9 \\
& Baseline - Set Transformer + LSTM & \textbf{140.9} & 33.9 & \textbf{88.7} & 17.7 & \textbf{49.2} & 24.5 & \textbf{38.6} & 18.2 & \textbf{33.3} & 11.2 & \textbf{35.0} & 12.8 \\
\cline{2-14}
& \textbf{Unmitigated Error (cm)} & 248.5 & 33.0 & 187.3 & 12.8 & 179.8 & 6.1 & 174.0 & 6.5 & 177.8 & 9.2 & 171.4 & 4.1 \\
& \textbf{Error Reduction (\gls{jaguard}) (\%)}\revg{\,$\uparrow$} & \multicolumn{2}{c}{43.2} & \multicolumn{2}{c}{52.8} & \multicolumn{2}{c}{72.5} & \multicolumn{2}{c}{77.8} & \multicolumn{2}{c}{81.1} & \multicolumn{2}{c}{79.3} \\
\midrule
\multirow{7}{*}{FM}
& Proposed - \gls{jaguard} & 108.0 & 26.2 & \textbf{54.8} & 14.9 & \textbf{38.9} & 11.9 & \textbf{29.7} & 12.7 & 32.3 & 6.1 & \textbf{41.5} & 8.8 \\
& SotA - GCNN~\cite{mohanty2023learning} & 108.8 & 26.3 & \textbf{54.4} & 13.7 & \textbf{38.9} & 12.0 & \textbf{29.7} & 14.0 & 31.5 & 8.6 & 46.7 & 9.7 \\
& Baseline - TSMixer & \textbf{102.8} & 24.5 & 57.9 & 11.3 & 43.3 & 6.7 & 36.0 & 10.0 & 33.9 & 8.3 & 44.4 & 10.3 \\
& Baseline - Set Transformer & 106.2 & 25.4 & 57.2 & 16.8 & \textbf{38.7} & 12.0 & \textbf{30.6} & 13.0 & 32.0 & 5.8 & \textbf{40.5} & 7.0 \\
& Baseline - Set Transformer + LSTM & 108.1 & 26.9 & \textbf{56.0} & 15.8 & \textbf{38.9} & 11.2 & \textbf{29.5} & 12.4 & \textbf{29.1} & 6.8 & \textbf{40.6} & 7.1 \\
\cline{2-14}
& \textbf{Unmitigated Error (cm)} & 224.9 & 19.6 & 189.4 & 13.3 & 178.8 & 6.5 & 175.1 & 3.7 & 175.6 & 9.3 & 163.2 & 9.6 \\
& \textbf{Error Reduction (\gls{jaguard}) (\%)}\revg{\,$\uparrow$} & \multicolumn{2}{c}{52.0} & \multicolumn{2}{c}{71.1} & \multicolumn{2}{c}{78.3} & \multicolumn{2}{c}{83.0} & \multicolumn{2}{c}{81.6} & \multicolumn{2}{c}{74.6} \\
\bottomrule
\end{tabular}
}
\label{tab:static}
\end{table*}

We evaluate \gls{jaguard} against TSMixer, Set Transformer, \revg{Set Transformer + LSTM}, and a purely spatial \gls{gcnn}, with results broken down by jamming type and power level \rev{at the held-out locations}.

Table~\ref{tab:static} presents the \rev{per-setting} results. Rows are organized by jamming type \rev{(CW, $3\times$CW, FM)}, with each method shown as a sub-row. Columns list jammer power levels, sweeping from $\mathbf{-45}$ to $\mathbf{-70}$\,dBm. For each jamming block, the final two rows report the unmitigated pre-correction error and the corresponding percentage error reduction achieved by the proposed model at that power level. \revg{\gls{jaguard} reduces the error by 43.2 to 83.0\% across the 18 settings, least at $-45$\,dBm (43.2 to 52.0\%, with uncorrected errors of 215.3 to 248.5\,cm). At rest, no model is best across the settings: \gls{jaguard} has the lowest MAE in 2 of the 18 settings and is not significantly above the lowest in 15, the Set Transformer + LSTM in 17 and the other baselines in 6 to 10 (bold in Table~\ref{tab:static}). Per setting, \gls{jaguard} is significantly below the GCNN, TSMixer and the Set Transformer in 7, 9 and 5 settings and above them in 0, 1 and 0, and it is below the Set Transformer + LSTM in 1 setting and above it in 2. The largest difference occurs under CW jamming at $-50$\,dBm, where the GCNN and TSMixer are 22.3 and 15.5\,cm above \gls{jaguard}.}

\rev{Averaged over the six held-out locations, \gls{jaguard} reduces the MAE from 174.9 to 56.9\,cm ($R=67.5\%$, Table~\ref{tab:summary}). The remaining error is largest at $-45$\,dBm, where the strongest jamming also disperses the position (Fig.~\ref{fig:variability}). At rest, all learned models with the identification head lie within 3.3\,cm of each other: the Set Transformer + LSTM is 0.4\,cm below \gls{jaguard} (95\% CI [0.1, 0.7], $p=0.026$), and the Set Transformer, TSMixer and \gls{gcnn} are \revg{0.2, 2.5 and 2.6\,cm above it (Table~\ref{tab:summary})}. Without its identification head, \gls{jaguard} reaches 59.8\,cm\revg{, 2.7\,cm (95\% CI [1.9, 3.6]) above the two-head model}.}

\rev{\label{rev:offset-results}Among the non-learned references, one mean offset of all jammed training recordings lowers the MAE to 67.9\,cm, and the mean offset of the true jamming setting, an oracle, to 55.0\,cm. \gls{jaguard}, which identifies the setting itself, comes within 1.9\,cm of this oracle. \revg{With the true jamming setting of the test window, \gls{jaguard} itself reaches 53.9\,cm at rest and 107.8\,cm on the routes, 3.0 and 12.0\,cm below its result with its own identification (Table~\ref{tab:summary}).} Because the correction is applied only to windows identified as jammed, the unjammed windows remain unchanged (\revg{false-positive rate} 0.0\%).}

\subsection{\revgsec{Kinematic Correction Along Known Routes}}
\label{sec:res:routes}

\begin{table*}[!ht]
\color{blue}
\centering
\setlength{\tabcolsep}{4pt}
\caption{Euclidean MAE (cm) on the held-out repetition of each route, for models trained on repetitions 1--3 of the same route. \revg{Bold: lowest MAE of the route, or of the mean over the routes, and the MAEs not significantly above it (paired 95\% bootstrap CI, Section~\ref{sec:metrics}).}}
\small
\begin{tabular}{l c ccccc}
\toprule
\textbf{Route} \revg{(MAE\,$\downarrow$)} & \textbf{Uncorr.} & \gls{jaguard} & GCNN & TSMixer & Set Tr. & Set Tr. + LSTM \\
\midrule
FE--UKC (bus) & 768.9 & \textbf{238.9} & 712.4 & 374.6 & 343.2 & 293.7 \\
Medvode--Topol (bus) & 804.5 & \textbf{257.1} & 729.4 & 340.5 & 347.7 & 312.6 \\
T01 Ljubljana \v{C}rnu\v{c}e--Dom\v{z}ale & 245.5 & \textbf{76.9} & 127.1 & 103.6 & 95.3 & 87.7 \\
T02 Dom\v{z}ale--Ljubljana \v{C}rnu\v{c}e & 220.2 & \textbf{85.4} & 120.0 & 105.9 & 95.2 & 103.8 \\
T03 Ptuj--Mo\v{s}kanjci & 281.9 & \textbf{90.9} & 105.9 & 98.1 & 95.8 & 94.9 \\
T04 Ljubljana--Ljubljana Zalog & 216.6 & \textbf{77.2} & 98.7 & 87.6 & 85.6 & \textbf{81.8} \\
T05 Grosuplje--\v{C}u\v{s}perk & 218.7 & \textbf{94.7} & 123.1 & 120.6 & 119.1 & 123.4 \\
T06 Ljubljana Zalog--Ljubljana & 207.9 & \textbf{103.8} & \textbf{111.7} & 115.7 & \textbf{102.8} & \textbf{102.8} \\
T07 \v{S}kofja Loka--Kranj & 284.9 & \textbf{99.8} & 174.6 & 137.0 & 118.9 & 130.0 \\
T08 Postojna--Prestranek & 216.9 & \textbf{109.4} & \textbf{118.7} & 121.9 & 113.5 & 119.0 \\
T09 Celje--\v{S}entjur & 216.1 & \textbf{89.7} & 106.5 & 99.8 & 97.5 & 96.2 \\
\midrule
Mean MAE & 334.7 & \textbf{120.3} & 229.8 & 155.0 & 146.8 & 140.6 \\
$R$ (\%)\revg{\,$\uparrow$} & -- & 64.0 & 31.3 & 53.7 & 56.1 & 58.0 \\
\bottomrule
\end{tabular}
\label{tab:routes}
\end{table*}

\rev{\label{rev:R2.1}Table~\ref{tab:routes} reports the MAE per route on the held-out repetition. \gls{jaguard} reduces the mean MAE over the eleven routes from 334.7 to 120.3\,cm ($R=64.0\%$) and has the lowest MAE of the five models on ten of the eleven routes. The strongest baseline, the Set Transformer + LSTM, is 20.2\,cm above \gls{jaguard} (95\% CI [16.4, 24.3], $p=2.3\times10^{-25}$, \gls{jaguard} lower on ten of the eleven routes). \revg{Per route, \gls{jaguard} is significantly below the GCNN, TSMixer, the Set Transformer and the Set Transformer + LSTM on 9, 11, 10 and 9 of the 11 routes and significantly above none of them on any route (bold in Table~\ref{tab:routes}).} Without its identification head, \gls{jaguard} reaches 123.1\,cm, which is not significantly different ($p=0.66$, Table~\ref{tab:summary}): the identification does not degrade the correction. \revg{The two bus routes have uncorrected errors of 768.9 and 804.5\,cm, roughly three times those of the railway routes (207.9 to 284.9\,cm). There, \gls{jaguard} leaves 238.9 and 257.1\,cm, whereas the GCNN, which has no receiver node, leaves 712.4 and 729.4\,cm (Table~\ref{tab:routes}).}}

\rev{\revg{Three elements thus contribute to the correction.} The identification selects the jamming setting, whose mean offset alone reaches 55.0\,cm at rest (oracle, Table~\ref{tab:summary})\revg{, but reduces the error on the routes by only 30.7\%}. In motion, the receiver inputs locate the error along the route: without the reported position, speed and heading, the error reduction on FE--UKC falls from 69.1 to 8.2\% (Section~\ref{sec:res:ablation}). \revg{Finally, the architecture separates the learned models in motion, with mean route MAEs from 120.3\,cm for \gls{jaguard} to 229.8\,cm for the GCNN (Table~\ref{tab:summary}).}}

\begin{table*}[!ht]
\color{blue}
\centering
\setlength{\tabcolsep}{4pt}
\caption{\revg{Mean Euclidean MAE (cm) over the held-out locations or routes, error reduction $R$ (\%), and paired MAE difference to \gls{jaguard} over the test recordings ($\Delta$MAE, cm, with 95\% bootstrap CI, positive when \gls{jaguard} is lower), which can differ from the difference of the MAE columns by up to 1.3\,cm. Oracles use the true jamming setting.}}
\small
\begin{tabular}{l ccc ccc}
\toprule
 & \multicolumn{3}{c}{\textbf{Static}} & \multicolumn{3}{c}{\textbf{\revg{Kinematic}}} \\
\cmidrule(lr){2-4}\cmidrule(lr){5-7}
\textbf{Method} & MAE\revg{\,$\downarrow$} & $R$\revg{\,$\uparrow$} & \revg{$\Delta$MAE [95\% CI]} & MAE\revg{\,$\downarrow$} & $R$\revg{\,$\uparrow$} & \revg{$\Delta$MAE [95\% CI]} \\
\midrule
Uncorrected & 174.9 & -- & -- & 334.7 & -- & -- \\
\midrule
\multicolumn{7}{l}{\textit{Models}} \\
\gls{jaguard}& 56.9 & 67.5 & -- & 120.3 & 64.0 & -- \\
GCNN~\cite{mohanty2023learning} & 59.8 & 65.8 & 2.6 [2.0, 3.4] & 229.8 & 31.3 & 108.6 [83.6, 134.5] \\
TSMixer & 59.6 & 65.9 & 2.5 [1.7, 3.2] & 155.0 & 53.7 & 33.4 [27.2, 39.9] \\
Set Transformer & 57.2 & 67.3 & 0.2 [$-$0.0, 0.5] & 146.8 & 56.1 & 26.0 [20.7, 31.7] \\
Set Transformer + LSTM & 56.5 & 67.7 & $-$0.4 [$-$0.7, $-$0.1] & 140.6 & 58.0 & 20.2 [16.4, 24.3] \\
\midrule
\multicolumn{7}{l}{\textit{Ablation}} \\
\gls{jaguard} without identification head & 59.8 & 65.8 & 2.7 [1.9, 3.6] & 123.1 & 63.2 & 2.6 [$-$0.5, 5.7] \\
\midrule
\multicolumn{7}{l}{\textit{References}} \\
Single offset of jammed recordings & 67.9 & 61.2 & 10.5 [9.1, 12.0] & 263.6 & 21.2 & 142.8 [117.9, 169.1] \\
Oracle: per-setting offset & 55.0 & 68.6 & $-$1.9 [$-$3.4, $-$0.5] & 231.8 & 30.7 & 110.7 [84.9, 138.8] \\
Oracle: \gls{jaguard}, true setting & 53.9 & 69.2 & $-$3.0 [$-$4.2, $-$1.8] & 107.8 & 67.8 & $-$12.0 [$-$18.1, $-$6.5] \\
\bottomrule
\end{tabular}
\label{tab:summary}
\end{table*}

\rev{\label{rev:rail}On held-out repetitions of the nine railway routes, \gls{jaguard} lowers the mean horizontal error of jammed fixes from 2.33 to 0.91\,m, and the 95th percentiles of the along-track and cross-track errors from 4.81 and 2.95\,m to 2.20 and 1.61\,m, below the 2.5\,m per axis at 95\% of the time that EUSPA requires for the absolute train position~\cite{euspa2025rail}. This requirement concerns the absolute safe train position, whose integrity the evaluation does not address (Section~\ref{sec:limitations}).}

\subsection{\revgsec{Calibration Along a New Route}}
\label{sec:res:deploy}

\begin{table}[!ht]
\color{blue}
\centering
\setlength{\tabcolsep}{4pt}
\begin{threeparttable}
\caption{\revg{Calibration along a new route}: error reduction $R$ (\%) on the held-out repetition of a model trained from scratch on $k$ recordings of the route ($k=0$: model of the ten other routes).}
\label{tab:calib}
\small
\begin{tabular}{l cc cc}
\toprule
 & \multicolumn{2}{c}{\textbf{FE--UKC (bus)}, \revg{$R$\,$\uparrow$}} & \multicolumn{2}{c}{\textbf{T07 (railway)}, \revg{$R$\,$\uparrow$}} \\
\cmidrule(lr){2-3}\cmidrule(lr){4-5}
$k$ & \gls{jaguard} & Set Tr. + LSTM & \gls{jaguard} & Set Tr. + LSTM \\
\midrule
0 & 3.3 & 1.8 & 33.5 & 27.5 \\
4 & 38.8 & 18.0 & 14.9 & 12.6 \\
10 & 56.2 & 31.7 & 47.3 & 37.1 \\
19 & 65.1 & 53.1 & 48.8 & 43.8 \\
38 & 68.1 & 58.4 & 60.9 & 52.1 \\
57 & 69.3 & 61.4 & 63.8 & 55.8 \\
57\revg{\tnote{$\dagger$}} & 68.9 & 61.8 & 65.0 & 54.3 \\
\bottomrule
\end{tabular}
\begin{tablenotes}\footnotesize
\item[\revg{$\dagger$}] \revg{With the validation repetition, as in Table~\ref{tab:routes}.}
\end{tablenotes}
\end{threeparttable}
\end{table}

\rev{To deploy \gls{jaguard} on a new route, the route is first recorded in the conducted setup, by simulating it, \revg{so that no jamming over the air is needed}. Table~\ref{tab:calib} reports the error reduction of a model trained on $k$ such recordings. \revg{Without recordings of the route ($k=0$), the model of the ten other routes reduces the error by only 3.3\% on the bus route FE--UKC and by 33.5\% on the railway route T07, so the correction transfers only partly to a new route.} Ten recordings, one without jamming and each jammer at three powers, raise the reduction from 3.3 to 56.2\% on FE--UKC and from 33.5 to 47.3\% on T07. With one recording per setting ($k=19$), FE--UKC reaches 65.1\%, against 68.9\% with all training repetitions, and with two per setting ($k=38$), T07 reaches 60.9\%, against 65.0\%. Four recordings with the jammers at $-45$\,dBm only, in contrast, give 14.9\% on T07\revg{, less than the 33.5\% of the model of the other routes}. \gls{jaguard} exceeds the Set Transformer + LSTM at every $k$ on both routes. With ten recordings, it already exceeds the reduction that the Set Transformer + LSTM reaches with 19 (56.2 vs.\ 53.1\% on FE--UKC and 47.3 vs.\ 43.8\% on T07), so it calibrates to a new route with about half the recordings.}

\subsection{\revsec{Jammers Left Out of Training}}
\label{sec:res:leftout}

\rev{\revg{A deployed model may also meet a jammer type or power that it has not seen in training.} Table~\ref{tab:leftout} reports the error reduction on a jammer type or power that was left out of training. A left-out $3\times$CW or FM jammer is still detected in all windows, and a left-out power in 98.0 to 100\% of them, whereas a left-out CW jammer is detected in only 77.7 to 80.3\%, consistent with the weaker effect of CW jamming, which leaves \revg{6.1 to 8.6 GPS satellites in the solution on average, against 8.2 to 8.4 without jamming}. The correction retains 0.93 to 0.99 of its reduction for a left-out $3\times$CW jammer and 0.70 to 0.94 for a left-out power, but only 0.49 to 0.59 for CW, and 0.90 at rest and 0.53 on the routes for FM. \revg{A jammer type missing from training is thus mostly detected, except CW jamming in about one fifth of the windows, but corrected less, in particular in motion.}}

\begin{table}[!t]
\color{blue}
\centering
\setlength{\tabcolsep}{2.4pt}
\caption{Jammer type or power left out of training: error reduction $R$ (\%) \revg{of \gls{jaguard}} on the left-out setting with and without it in training, their ratio, and the detection rate (\%) of the model trained without it.}
\begin{tabular}{l cccc cccc}
\toprule
 & \multicolumn{4}{c}{\textbf{Static}} & \multicolumn{4}{c}{\textbf{\revg{Kinematic}}} \\
\cmidrule(lr){2-5}\cmidrule(lr){6-9}
 & \multicolumn{2}{c}{\revg{$R$ (\%)\,$\uparrow$}} & & & \multicolumn{2}{c}{\revg{$R$ (\%)\,$\uparrow$}} & & \\
\cmidrule(lr){2-3}\cmidrule(lr){6-7}
\textbf{Left out} & With & W/o & Ratio\revg{\,$\uparrow$} & Det.\revg{\,$\uparrow$} & With & W/o & Ratio\revg{\,$\uparrow$} & Det.\revg{\,$\uparrow$} \\
\midrule
CW & 64.2 & 37.8 & 0.59 & 80.3 & 59.4 & 29.0 & 0.49 & 77.7 \\
3$\times$CW & 64.5 & 63.6 & 0.99 & 100.0 & 66.0 & 61.2 & 0.93 & 100.0 \\
FM & 72.4 & 64.9 & 0.90 & 100.0 & 69.0 & 36.9 & 0.53 & 100.0 \\
$-45$\,dBm, all types & 46.9 & 43.8 & 0.93 & 100.0 & 49.8 & 41.8 & 0.84 & 100.0 \\
$-70$\,dBm, all types & 74.5 & 69.8 & 0.94 & 99.4 & 61.1 & 43.0 & 0.70 & 98.0 \\
\bottomrule
\end{tabular}
\label{tab:leftout}
\end{table}

\subsection{\revsec{Satellites of All Constellations}}
\label{sec:res:gec}

\begin{table}[!t]
\color{blue}
\centering
\setlength{\tabcolsep}{2.5pt}
\caption{Two-head models with the satellites of GPS only (main results) or of all constellations (GPS, Galileo, BeiDou) as nodes, on the same windows: MAE (cm) and exact jamming setting identified (\%).}
\small
\begin{tabular}{ll cc cc}
\toprule
 & & \multicolumn{2}{c}{\textbf{MAE}\revg{\,$\downarrow$}} & \multicolumn{2}{c}{\textbf{Exact setting}\revg{\,$\uparrow$}} \\
\cmidrule(lr){3-4}\cmidrule(lr){5-6}
\textbf{Model} & \textbf{Sat.} & Static & \revg{Kinematic} & Static & \revg{Kinematic} \\
\midrule
\gls{jaguard} & GPS & 56.9 & 120.3 & 63.5 & 60.1 \\
\gls{jaguard} & all & 54.5 & 112.1 & 95.4 & 93.2 \\
Set Tr. + LSTM & GPS & 56.5 & 140.6 & 65.7 & 61.0 \\
Set Tr. + LSTM & all & 54.4 & 137.4 & 96.7 & 93.5 \\
\bottomrule
\end{tabular}
\label{tab:gec}
\end{table}

\begin{table*}[!t]
\color{blue}
\centering
\setlength{\tabcolsep}{3pt}
\caption{\revg{Ablation of \gls{jaguard} at rest and on the route FE--UKC: error reduction $R$ (\%), paired MAE difference to the proposed model ($\Delta$MAE, cm, mean and 95\% bootstrap CI, positive when the variant is worse), and type and exact-setting accuracy (\%).}}
\small
\begin{tabular}{l cccc cccc}
\toprule
 & \multicolumn{4}{c}{\textbf{Static}} & \multicolumn{4}{c}{\textbf{\revg{Kinematic, FE--UKC}}} \\
\cmidrule(lr){2-5}\cmidrule(lr){6-9}
\textbf{Variant} & $R$\revg{\,$\uparrow$} & $\Delta$MAE [95\% CI] & Type\revg{\,$\uparrow$} & Exact\revg{\,$\uparrow$} & $R$\revg{\,$\uparrow$} & $\Delta$MAE [95\% CI] & Type\revg{\,$\uparrow$} & Exact\revg{\,$\uparrow$} \\
\midrule
Proposed \revg{($W=10$, hidden size 64)} & 68.2 & -- & 97.5 & 64.6 & 69.1 & -- & 97.0 & 57.9 \\
\midrule
\multicolumn{9}{l}{\revg{\textit{Inputs}}} \\
w/o SNR & 62.3 & 10.4 [8.8, 12.0] & 83.7 & 27.3 & 61.4 & 62.7 [26.7, 104.6] & 84.0 & 28.3 \\
w/o elevation & 68.1 & 0.3 [0.1, 0.5] & 97.2 & 64.3 & 69.1 & $-$0.0 [$-$3.7, 3.4] & 97.5 & 57.7 \\
w/o azimuth & 68.2 & 0.1 [$-$0.1, 0.2] & 97.1 & 64.0 & 68.8 & 2.5 [$-$1.4, 6.4] & 95.5 & 55.4 \\
w/o SNR, azimuth, elevation & 60.0 & 14.5 [12.6, 16.4] & 73.6 & 19.8 & 56.6 & 102.0 [42.2, 170.9] & 62.1 & 15.9 \\
With receiver coordinates & 67.4 & 1.4 [0.8, 2.1] & 95.3 & 61.8 & \multicolumn{4}{c}{--} \\
Receiver position only & \multicolumn{4}{c}{--} & 59.8 & 75.9 [64.7, 86.9] & 96.4 & 56.2 \\
Receiver speed and heading only & \multicolumn{4}{c}{--} & 38.9 & 246.1 [196.9, 289.8] & 96.5 & 57.6 \\
w/o receiver inputs & \multicolumn{4}{c}{--} & 8.2 & 496.3 [431.0, 550.4] & 96.7 & 57.5 \\
\midrule
\multicolumn{9}{l}{\revg{\textit{Graph}}} \\
w/o edges ($W=10$) & 61.1 & 12.6 [11.0, 14.3] & 35.9 & 6.0 & 65.8 & 26.9 [1.9, 55.9] & 35.8 & 6.2 \\
w/o history ($W=1$) & 61.1 & 12.7 [11.1, 14.3] & 35.9 & 6.0 & 53.9 & 123.6 [98.5, 144.5] & 37.4 & 6.2 \\
Satellite mesh, no receiver ($W=1$) & 62.2 & 10.5 [8.8, 12.2] & 68.0 & 31.5 & 6.7 & 508.7 [442.1, 562.9] & 65.9 & 30.3 \\
Satellite mesh, no receiver ($W=10$) & 68.3 & $-$0.1 [$-$0.3, 0.1] & 98.5 & 66.0 & 8.2 & 497.0 [432.4, 550.7] & 98.6 & 60.5 \\
Satellite$\to$receiver edges only & 68.1 & 0.2 [$-$0.0, 0.4] & 97.3 & 61.8 & 70.1 & $-$8.2 [$-$12.8, $-$3.7] & 96.2 & 55.8 \\
With inter-satellite edges & 68.3 & $-$0.1 [$-$0.2, 0.0] & 97.7 & 64.7 & 68.3 & 6.0 [0.2, 12.7] & 96.7 & 57.2 \\
Attention aggregation (GATv2) & 68.2 & 0.0 [$-$0.2, 0.2] & 96.9 & 65.4 & 68.6 & 4.0 [$-$3.3, 12.3] & 96.4 & 57.2 \\
\midrule
\multicolumn{9}{l}{\revg{\textit{Memory}}} \\
Memory reset during outages & 68.3 & $-$0.1 [$-$0.3, 0.0] & 97.6 & 61.7 & 69.7 & $-$5.2 [$-$11.0, 0.8] & 97.8 & 59.3 \\
One memory shared by all satellites & 68.1 & 0.3 [0.1, 0.5] & 93.9 & 60.3 & 70.1 & $-$8.3 [$-$15.5, $-$1.2] & 89.3 & 51.4 \\
\midrule
\multicolumn{9}{l}{\revg{\textit{Size}}} \\
Window 5 & 67.8 & 0.8 [0.5, 1.2] & 95.2 & 56.2 & 66.2 & 23.8 [11.2, 34.0] & 95.4 & 55.6 \\
Window 20 & 68.4 & $-$0.3 [$-$0.5, $-$0.0] & 98.4 & 69.5 & 69.0 & 0.4 [$-$8.4, 11.2] & 97.3 & 58.4 \\
Window 30 & 68.4 & $-$0.4 [$-$0.6, $-$0.1] & 98.2 & 68.1 & 68.5 & 4.7 [$-$5.8, 17.3] & 96.3 & 57.3 \\
Hidden size 16 & 67.9 & 0.7 [0.3, 1.0] & 97.6 & 62.5 & 65.7 & 27.3 [19.9, 34.3] & 91.9 & 50.1 \\
Hidden size 32 & 68.2 & 0.1 [$-$0.1, 0.2] & 97.8 & 64.5 & 68.8 & 2.4 [$-$2.0, 7.2] & 96.7 & 56.4 \\
Hidden size 128 & 68.1 & 0.3 [0.2, 0.4] & 97.4 & 63.9 & 68.0 & 8.8 [3.5, 14.4] & 97.0 & 57.4 \\
Hidden size 256 & 68.1 & 0.2 [0.0, 0.3] & 96.9 & 61.8 & 66.8 & 18.5 [9.3, 29.0] & 96.2 & 57.0 \\
\bottomrule
\end{tabular}
\label{tab:ablation}
\end{table*}

\rev{\label{rev:R2.4-results}Table~\ref{tab:gec} repeats the evaluation with the GPS, Galileo and BeiDou satellites as nodes, on the same windows. Each satellite node also carries its constellation and whether the receiver uses it in its solution, which the NMEA GSA sentence reports. \gls{jaguard} then lowers the MAE from 56.9 to 54.5\,cm at rest and from 120.3 to 112.1\,cm on the routes, and identifies the exact jamming setting in 95.4 and 93.2\% of the windows, against 63.5 and 60.1\% with GPS satellites only. With the same inputs, the Set Transformer + LSTM reaches 54.4\,cm at rest\revg{, 0.1\,cm below \gls{jaguard},} and 137.4\,cm on the routes, 25.3\,cm above \gls{jaguard}\revg{, and identifies the exact setting slightly more often (96.7 and 93.5\%)}.}

\subsection{Ablation Study}
\label{sec:res:ablation}

\begin{table}[!t]
\color{blue}
\centering
\setlength{\tabcolsep}{3.5pt}
\caption{\revg{Inference cost on a Raspberry Pi~5 per 10-s window: mean latency at rest and in motion, and 99th percentile in motion.}}
\begin{tabular}{l rr rrr}
\toprule
 & & & \multicolumn{3}{c}{\revg{Latency (ms)\,$\downarrow$}} \\
\cmidrule(lr){4-6}
\textbf{Model} & Param.\revg{\,$\downarrow$} & MFLOP\revg{\,$\downarrow$} & Rest & Motion & p99 \\
\midrule
\gls{jaguard} & 69,205 & 5.8 & 37.3 & 37.7 & 38.9 \\
GCNN~\cite{mohanty2023learning} & 5,111 & 0.1 & 1.2 & 1.1 & 1.2 \\
TSMixer & 47,050 & 0.7 & 1.4 & 1.4 & 1.4 \\
Set Transformer & 90,453 & 13.3 & 6.1 & 6.2 & 7.6 \\
Set Tr. + LSTM & 89,941 & 9.2 & 4.2 & 4.3 & 4.8 \\
\gls{jaguard}, GATv2 & 70,229 & 8.6 & 65.2 & 64.3 & 66.0 \\
\bottomrule
\end{tabular}
\label{tab:computation}
\end{table}

\rev{\label{rev:R1.7}Table~\ref{tab:ablation} lists the variants \revg{by group, against an MAE of the proposed model of 55.9\,cm at rest and 252.3\,cm on the route, on the windows of the ablation (Section~\ref{sec:ablation})}.}

\rev{\revg{\textit{Inputs.}} The \gls{snr} carries the identification: without it, the jammer-type accuracy falls from 97.5 to 83.7\% at rest and from 97.0 to 84.0\% on the route, the \revg{false-positive rate} at rest rises from 0.0 to 38.0\%, and the MAE rises by 10.4\,cm at rest and by 62.7\,cm on the route. Removing the azimuth or the elevation changes the MAE by at most 0.3\,cm at rest and 2.5\,cm on the route. Each jammer type leaves its own C/N$_0$ pattern on the GPS satellites, measured at the six static locations against the unjammed recordings at the same time. CW jamming lowers the C/N$_0$ of all satellites by 3.5 to 17.5\,dB, depending on the power, with less than 1\,dB difference between satellites. $3\times$CW jamming removes 71 to 83\% of the satellites at every power. FM jamming lowers the C/N$_0$ by up to 24.7\,dB, with up to 10.3\,dB difference between satellites. \revg{The receiver inputs} carry the correction in motion: without them, \gls{jaguard} reduces the error on the route by only 8.2\%, against 69.1\%\revg{, with the position only by 59.8\%, and with the speed and heading only by 38.9\%}. At rest, adding the receiver coordinates raises the MAE by 1.4\,cm.}

\rev{\revg{\textit{Graph.}} \revg{The receiver node receives the satellite information through the satellite states of the previous epoch (Eq.~\ref{eq:gate_activation}). Without edges, or with a single epoch ($W=1$), this information therefore does not reach it.} At rest, the edgeless or single epoch model identifies the jammer type in only 35.9\% of the windows, and its correction reduces to a constant offset \revg{($R=61.1\%$, against 61.2\% for a single offset, Table~\ref{tab:summary})}. The satellite mesh without a receiver node, as in~\cite{mohanty2023learning}, reaches the error of \gls{jaguard} at rest with ten epochs \revg{(10.5\,cm above it with one epoch)}, but reduces the error on the route by only 8.2\%, because it has no node for the receiver inputs. The edges from the receiver to the satellites raise the exact-setting accuracy at rest from 61.8 to 64.6\% \revg{and change the MAE by only 0.2\,cm, whereas on the route the variant without them, with satellite-to-receiver edges only, is 8.2\,cm below the proposed model. Adding inter-satellite edges leaves the MAE unchanged at rest and raises it by 6.0\,cm on the route.} \label{rev:R1.4-results}Attention-based aggregation (GATv2) leaves the MAE unchanged and increases the latency by a factor of 1.7 (Table~\ref{tab:computation}).}

\rev{\revg{\textit{Memory.}} The separate memory per satellite and its freeze during outages improve the identification. Compared with one memory shared by all satellites, the separate memories identify the jammer type in 97.5 vs.\ 93.9\% of the windows at rest and in 97.0 vs.\ 89.3\% on the route, and the exact setting in 64.6 vs.\ 60.3\% and 57.9 vs.\ 51.4\%. Compared with a reset during outages, the freeze raises the exact-setting accuracy at rest from 61.7 to 64.6\% (paired difference 3.2 percentage points, 95\% CI [2.0, 4.6], higher at each of the six held-out locations). The gain occurs under $3\times$CW and FM jamming (3.7 and 4.8 points), under which 85.6 and 33.8\% of the GPS satellites \revg{tracked at the end of the window had lost lock} within the preceding 10\,s, and not under CW (1.2 points, 95\% CI $[-0.3, 2.4]$), under which 4.3\% did. On the route, the freeze does not change the identification ($-1.3$ points, 95\% CI $[-4.7, 2.2]$). \revg{Neither the separate memory nor the freeze lowers the MAE by more than 0.3\,cm at rest, and on the route the shared memory is 8.3\,cm below the proposed model.}}

\rev{\revg{\textit{Size.}} \label{rev:R1.6-results}A window of 5\,s and a hidden size of 16 raise the route MAE by 23.8 and 27.3\,cm, whereas longer windows and larger hidden sizes do not lower it\revg{. At rest, windows of 20 and 30\,s lower the MAE by 0.3 and 0.4\,cm and raise the exact-setting accuracy from 64.6 to 69.5 and 68.1\%, which the route does not show. The route thus determines the choice of} $W=10$ and a hidden size of 64.}

\revg{In summary, the message passing from the satellites to the receiver, the temporal recurrence and, in motion, the receiver inputs are the primary drivers of position correction. The separate memory per satellite, its freeze and the bidirectional receiver-to-satellite edges help with fine-grained jamming identification, accepting a statistically negligible or minor (<3.3\%) position error trade-off in motion to ensure robust multi-task operation.}

\subsection{Computational Complexity Evaluation}
\label{sec:res:computation}

To ensure continuous real-time positioning, mitigation models
must complete inference within the \gls{gnss} receiver's 1\,Hz
sampling epoch. \rev{On the Raspberry Pi~5 (Table~\ref{tab:computation}), \gls{jaguard} requires 37.3\,ms per window at rest and 37.7\,ms in motion (99th percentile 38.9\,ms\revg{, standard deviation at most 1.5\,ms for all models}), less than 4\% of the 1-s epoch. It is the slowest of the five models, although it needs fewer operations than the two attention-based models, because its graph convolution runs inside the recurrence as ten sequential steps of small sparse operations, whereas the attention-based models process the whole window in parallel dense operations.}

\section{Operational Boundaries and Limitations}
\label{sec:limitations}

\gls{jaguard} operates in the position domain rather than correcting raw measurements. This is a deliberate design choice: because jamming degrades all satellites simultaneously, traditional integrity logic, built to exclude isolated faults, would often discard every measurement and terminate navigation. By correcting the final position, \gls{jaguard} captures the aggregate distortion while keeping the receiver producing a solution. The tradeoff is that the receiver must still output a valid (though biased) fix. Under complete tracking loss, no position is available and \gls{jaguard} cannot operate. \rev{\label{rev:R2.5}Complete denial is, however, explicit: the receiver stops reporting a fix, and the dependent system can switch to other sensors. The case that \gls{jaguard} addresses is the silent one, in which the receiver keeps reporting a position without a flag, as under the FM jamming in our recordings (\revg{Section~\ref{sec:res:ident}}).}

\rev{\label{rev:R2.6-limits}\revg{In the static scenario}, the correction transfers to held-out locations. \revg{In the kinematic scenario}, it is learned for each route from recordings of that route, as for the repeated trips of trains and buses. A deployment on a new route therefore calibrates the model with recordings of that route (Section~\ref{sec:res:deploy}). \label{rev:field-limits}Over the air, laboratory training sufficed to detect strong interference, and interference with a C/N$_0$ drop of 3 to 6\,dB was detected in \revg{23.2\% of the Jammertest windows and 25.0\%} of the cruise events (Table~\ref{tab:field}). The correction is not evaluated over the air, because neither field dataset provides a reference position.}

\rev{The recordings are conducted, not over the air, so they contain no multipath, obstruction or antenna effects, and they cover one receiver model and one simulated date and time of day. \revg{Recording our dataset over the air would require transmitting jamming signals, which is illegal in most countries outside authorized events such as Jammertest~\cite{gerrard2026jammertest}.} The evaluation addresses the accuracy of the corrected position, not its integrity. \label{rev:R1.8}\revg{For over-the-air recordings, the reference position would have to come from carrier-phase relative positioning against a reference station or a network of continuously operating reference stations~\cite{kaplan2017gnss}: static relative positioning in the static scenario, and real-time (RTK) or post-processed (PPK) kinematic processing in the kinematic scenario.} \label{rev:R1.9}The released UBX logs also contain the receiver's code-phase and Doppler measurements (UBX-RXM-MEASX), which allow the interference to be analyzed at the measurement level. The receiver does not output carrier-phase measurements~\cite{ublox_m10_ifd}.}

\section{Conclusions}
\label{sec:conclusions}

In this work, we introduced \gls{jaguard}, a deep temporal graph network that addresses a resilient \gls{gnss} positioning challenge by formulating \gls{gnss} jamming mitigation as a dynamic graph \revg{identification and} regression problem. By modeling the satellite constellation as a heterogeneous graph, \gls{jaguard} inherently adapts to signal volatility and intermittent outages without requiring artificial data imputation. \rev{It identifies the interference jointly with the correction and applies the correction only to fixes identified as jammed.}

\revg{On hardware-in-the-loop recordings of a commercial multi-constellation receiver \revg{in a static scenario} at six locations and \revg{in a kinematic scenario} on eleven routes, \gls{jaguard} detected 99.8 to 99.9\% of the jammed windows, including the FM jamming that the receiver internal monitor detects in 3.7 to 6.9\% of the windows, and reduced the error of the reported position by 67.5\% at held-out \revg{static} locations and by 64.0\% on held-out repetitions of the \revg{kinematic} routes, \revg{where it was} 20.2\,cm below the strongest baseline. On a new route, ten calibration recordings raised its reduction to 47.3 to 56.2\%\revg{, against 3.3 to 33.5\% without them}, more than the strongest baseline reached with 19.}

\revg{Trained only on laboratory data, \gls{jaguard} detected interference in over-the-air recordings of other receivers: 97.7\% of the Jammertest 2024 windows with a C/N$_0$ drop of at least 10\,dB and all 28 interference events with a drop above 6\,dB on an Antarctic cruise, without false positives on the clean Jammertest windows, for which TSMixer and the Set Transformer had false-positive rates of 60.6 and 8.8\%.} \revg{For a model trained without one of the three jammer types, the omitted type was detected in 100.0\% of the windows for $3\times$CW and FM jamming and in 77.7 to 80.3\% for CW jamming, whereas the error reduction decreased to 0.49 to 0.59 of that obtained with the type included in training for CW jamming and, in the kinematic scenario, for FM jamming.} \revg{In the static scenario, the MAE of the attention-based baselines was within 0.4\,cm of that of \gls{jaguard}, which therefore achieved a significantly lower MAE than these baselines only in the kinematic scenario.}

\section*{Acknowledgements}

This work was supported by the Slovenian Research Agency (P2-0016).

%\section*{Author contributions statement}

%B.B. and I.K. conceived the experiments,  I.K. conducted the experiments, B.B. and I.K. analysed the results. I.K. and B.B. wrote the original draft. All authors reviewed the manuscript.

\bibliographystyle{IEEEtran}
\bibliography{sampletmc}

\begin{IEEEbiography}[{\includegraphics[
width=1in,height=1.25in,keepaspectratio
]{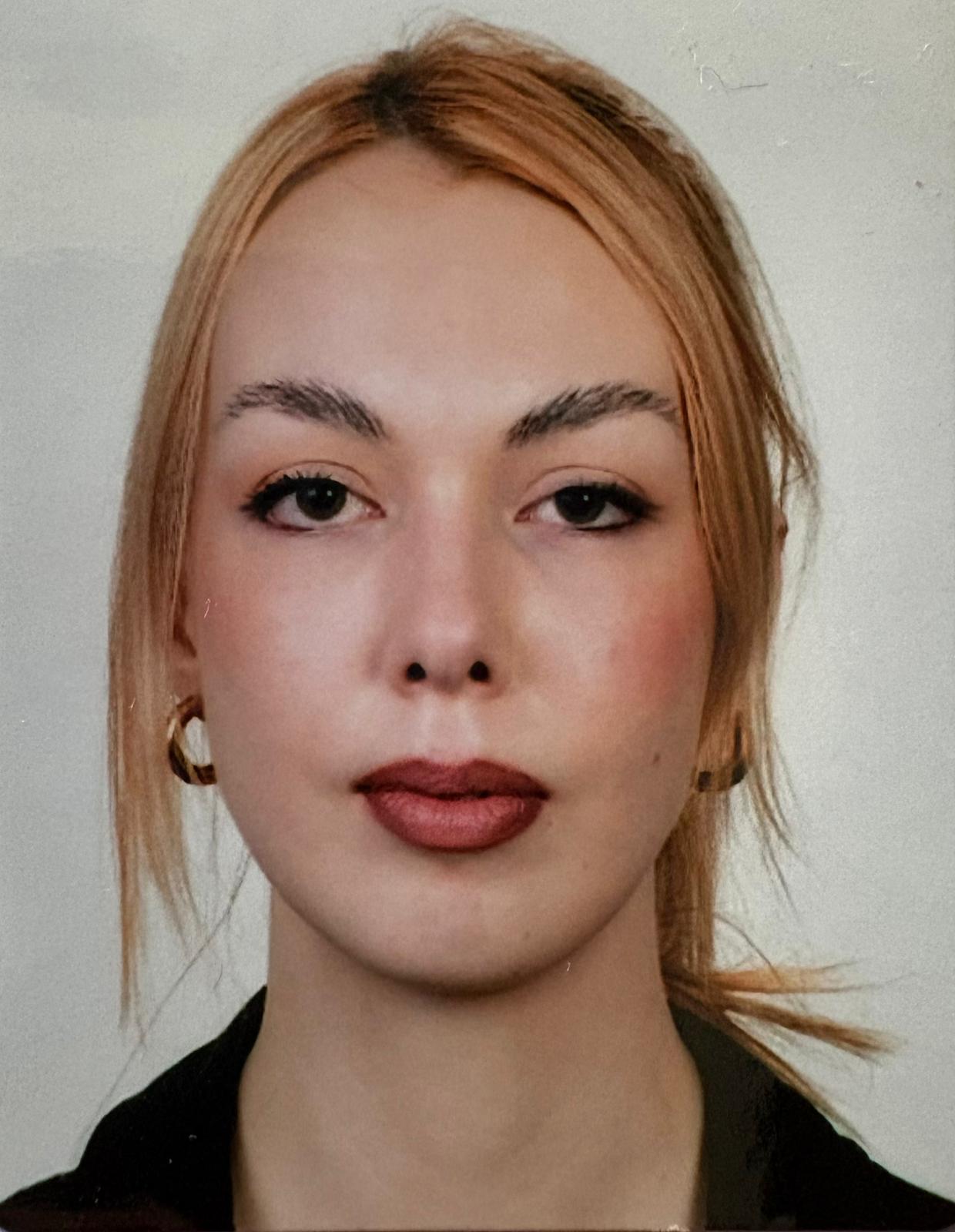}}]{Ivana Kesi\'c}
received the M.Sc. degree in Electrical Engineering
from the University of Ljubljana, Slovenia, in 2026.
She is currently pursuing the Ph.D. degree in
Information and Communication Technologies at the
Jo\v{z}ef Stefan International Postgraduate School.
Since 2023, she has been a member of SensorLab at
the Jo\v{z}ef Stefan Institute. Her research interests
include time-series analysis, distribution shift
detection, and graph neural networks, with
applications in communication networks and GNSS.
\end{IEEEbiography}

\begin{IEEEbiography}[{\includegraphics[
width=1in,height=1.25in,keepaspectratio
]{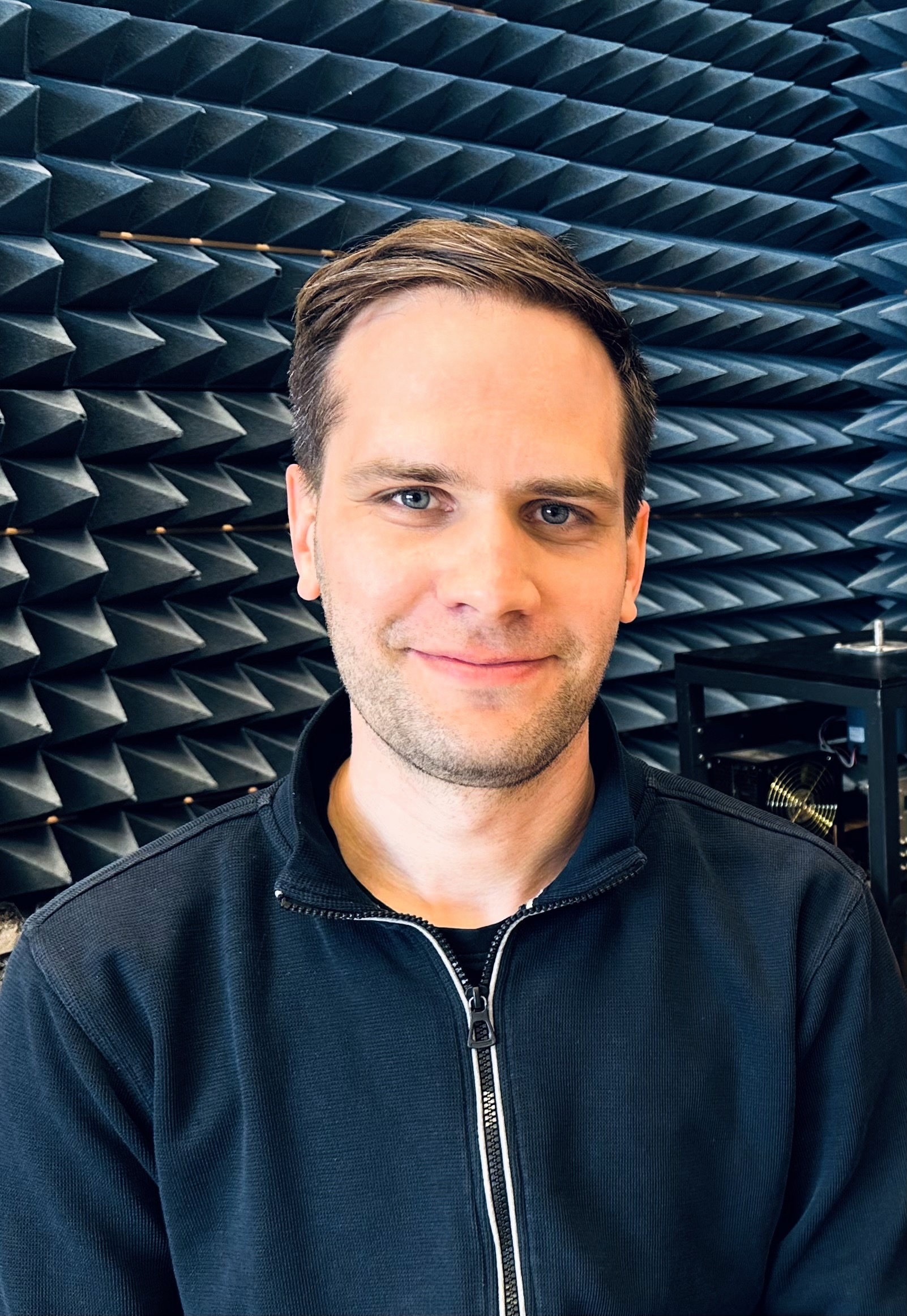}}]{Alja\v{z} Blatnik}
received the Ph.D. degree in Electrical Engineering
from the University of Ljubljana, Slovenia, in 2023.
Since 2017, he has been with the Faculty of Electrical
Engineering, University of Ljubljana, where he is
involved in teaching and research. His research
interests include radio-frequency engineering,
wireless and satellite communications, radar systems,
antenna structures, and advanced signal processing.
\end{IEEEbiography}

\begin{IEEEbiography}[{\includegraphics[
width=1in,height=1.25in,keepaspectratio
]{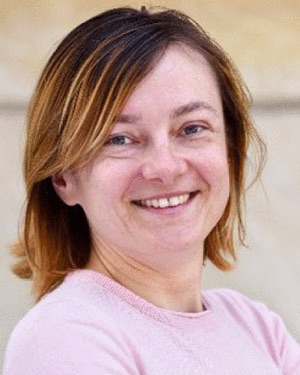}}]{Carolina Fortuna}
has more than 15 years of experience in artificial
intelligence and communication networks. She is
currently a Research Associate Professor at the
Jo\v{z}ef Stefan Institute, Slovenia, where she leads
SensorLab. Previously, she was a postdoctoral
researcher at Ghent University, Belgium, and a
visiting researcher at Stanford University, USA.
She has led several EU-funded research projects
and has coauthored more than 100 peer-reviewed
publications in leading international journals
and conferences. Her research interests include
artificial intelligence, communication networks,
and next-generation smart infrastructures.
\end{IEEEbiography}

\begin{IEEEbiography}[{\includegraphics[
width=1in,height=1.25in,keepaspectratio
]{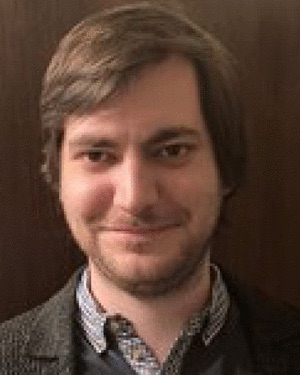}}]{Bla\v{z} Bertalani\v{c}}
(Member, IEEE) received the Ph.D. degree with
highest distinction from the Faculty of Electrical
Engineering, University of Ljubljana, Slovenia.
He is currently a researcher with SensorLab,
Jo\v{z}ef Stefan Institute. His research interests
include the development of machine learning and
artificial intelligence algorithms, particularly
for time-series analysis and smart infrastructures.
He also holds several leadership positions within
the Slovenian IEEE Chapter and has authored more
than 15 IEEE publications.
\end{IEEEbiography}

\vfill 

\end{document}